\documentclass[10pt]{article} 

\usepackage[preprint]{rlj} 

\usepackage{amssymb}            
\usepackage{mathtools}          
\usepackage{mathrsfs}           
\usepackage{graphicx}           
\usepackage{subcaption}         
\usepackage[space]{grffile}     
\usepackage{url}                
\usepackage{lipsum}    
\usepackage{wrapfig}
\usepackage{dsfont}
\usepackage{booktabs}
\usepackage{multirow}
\usepackage{bm} 
\usepackage{rotating} 

\usepackage{relsize}
\usepackage{svg}

\usepackage{booktabs}   
\usepackage{colortbl}   
\usepackage[frozencache, cachedir=.]{minted}

\usepackage[normalem]{ulem} 
\renewcommand{\underline}{\uline}

\tcbset{on line, 
        boxsep=3pt, left=0pt, right=0pt, top=0pt, bottom=0pt, boxrule=0pt,
        colframe=white, colback=white
}

\definecolor{Orange}{HTML}{FF6F20}  
\definecolor{Blue}{HTML}{0072B8}    

\colorlet{LightOrange}{Orange!40!white}  
\colorlet{LightBlue}{Blue!40!white}      

\newcommand{\ci}[1]{\scriptstyle\ [#1] }

\title{Deep Reinforcement Learning via Object-Centric \\ Attention}

\setrunningtitle{DRL via Object-Centric Attention}

\author{Jannis Blüml\textsuperscript{1,2,$\ast$}, Cedric Derstroff\textsuperscript{1,2,$\ast$}, Bjarne Gregori\textsuperscript{1}, Elisabeth Dillies\textsuperscript{3}, Quentin Delfosse\textsuperscript{1}, Kristian Kersting\textsuperscript{1,2,4,5}}

\emails{blueml@cs.tu-darmstadt.de \ cedric.derstroff@tu-darmstadt.de}

\affiliations{
$^{1}$\textbf{Department of Computer Science, Technical University of Darmstadt, Germany}\\
$^{2}$\textbf{Hessian Center for Artificial Intelligence (hessian.AI), Germany}\\
$^{3}$\textbf{Sorbonne Université, Paris, France }\\
$^{4}$\textbf{German Research Center for Artificial Intelligence (DFKI), Germany}\\
$^{5}$\textbf{Centre for Cognitive Science, Darmstadt, Germany}
\par 
$^\ast$ indicating equal contribution
}

\contribution{
    We propose Object-Centric Attention via Masking (OCCAM), a novel abstraction method that enhances RL generalization by selectively filtering background information while preserving task-relevant entities.
}
{
    Prior work on object-centric RL has primarily focused on symbolic reasoning or learned slot-based attention, often requiring task-specific priors or complex object extraction pipelines.
}

\contribution{
    We demonstrate that OCCAM improves robustness to visual perturbations while maintaining or surpassing performance compared to standard pixel-based RL approaches.
}
{
    These improvements are observed in controlled Atari experiments with perturbations from the HackAtari environment but have not yet been tested on real-world RL tasks or continuous control benchmarks.
}

\contribution{
    We show that OCCAM can reduce shortcut learning in RL agents, mitigating reliance on spurious correlations without requiring explicit symbolic representations.
}
{
    \citet{delfosse2024interpretable} suggested that symbolic reasoning is necessary to resolve shortcut learning issues in RL; our findings indicate that structured abstraction alone can achieve similar effects in specific environments.
}


\summary{
Deep RL agents trained on raw pixel inputs often struggle to generalize beyond their training environments, relying on spurious correlations and background artifacts that do not contribute to decision-making. Object-centric representations offer a promising alternative, but existing approaches typically require task-specific assumptions, pretraining, or additional supervision, limiting their scalability and adaptability.

Object-Centric Attention via Masking (OCCAM) is a simple yet effective framework that selectively preserves object entities while filtering out non-object pixel details. Inspired by cognitive science and Occam’s Razor, OCCAM enforces a structured inductive bias without relying on explicit symbolic reasoning or complex object extraction pipelines. By applying a lightweight masking strategy, OCCAM refines visual input representations, improving robustness against visual perturbations while having comparable or superior performance to conventional deep RL agents.

Empirical evaluations on Atari benchmarks demonstrate that OCCAM significantly mitigates the reliance on spurious correlations while maintaining task-agnostic applicability across different environments. Our findings suggest that object-centric attention can serve as a practical middle ground between raw pixel inputs and entirely symbolic representations by using objects to support and improve visual reasoning.

}

\begin{document}

\maketitle  

\begin{abstract}
Deep reinforcement learning agents, trained on raw pixel inputs, often fail to generalize beyond their training environments, relying on spurious correlations and irrelevant background details. To address this issue, object-centric agents have recently emerged. However, they require different representations tailored to the task specifications. Contrary to deep agents, no single object-centric architecture can be applied to any environment. 
Inspired by principles of cognitive science and Occam’s Razor, we introduce Object-Centric Attention via Masking (OCCAM), which selectively preserves task-relevant entities while filtering out irrelevant visual information. 
Specifically, OCCAM takes advantage of the object-centric inductive bias.
Empirical evaluations on Atari benchmarks demonstrate that OCCAM significantly improves robustness to novel perturbations and reduces sample complexity while showing similar or improved performance compared to conventional pixel-based RL. These results suggest that structured abstraction can enhance generalization without requiring explicit symbolic representations or domain-specific object extraction pipelines.
\end{abstract}

\section{Introduction}
\label{sec:introduction}



Human visual processing, akin to the dual-system theory of fast and slow thinking \citep{kahneman2011thinking}, operates in two phases: a rapid, automatic process that scans the visual field to detect salient features \citep{treisman1985preattentive} and a sequential, focused attention mechanism that extracts complex representations from localized regions \citep{treisman1980}. This hierarchical approach highlights the role of abstract representations—where key entities and their relationships serve as fundamental building blocks for human reasoning and planning \citep{baars1993cognitive, baars2002conscious, Bengio17Prior, goyal2022inductive}.

\begin{wrapfigure}[22]{r}{0.5\linewidth}
    \centering
    \includegraphics[width=\linewidth]{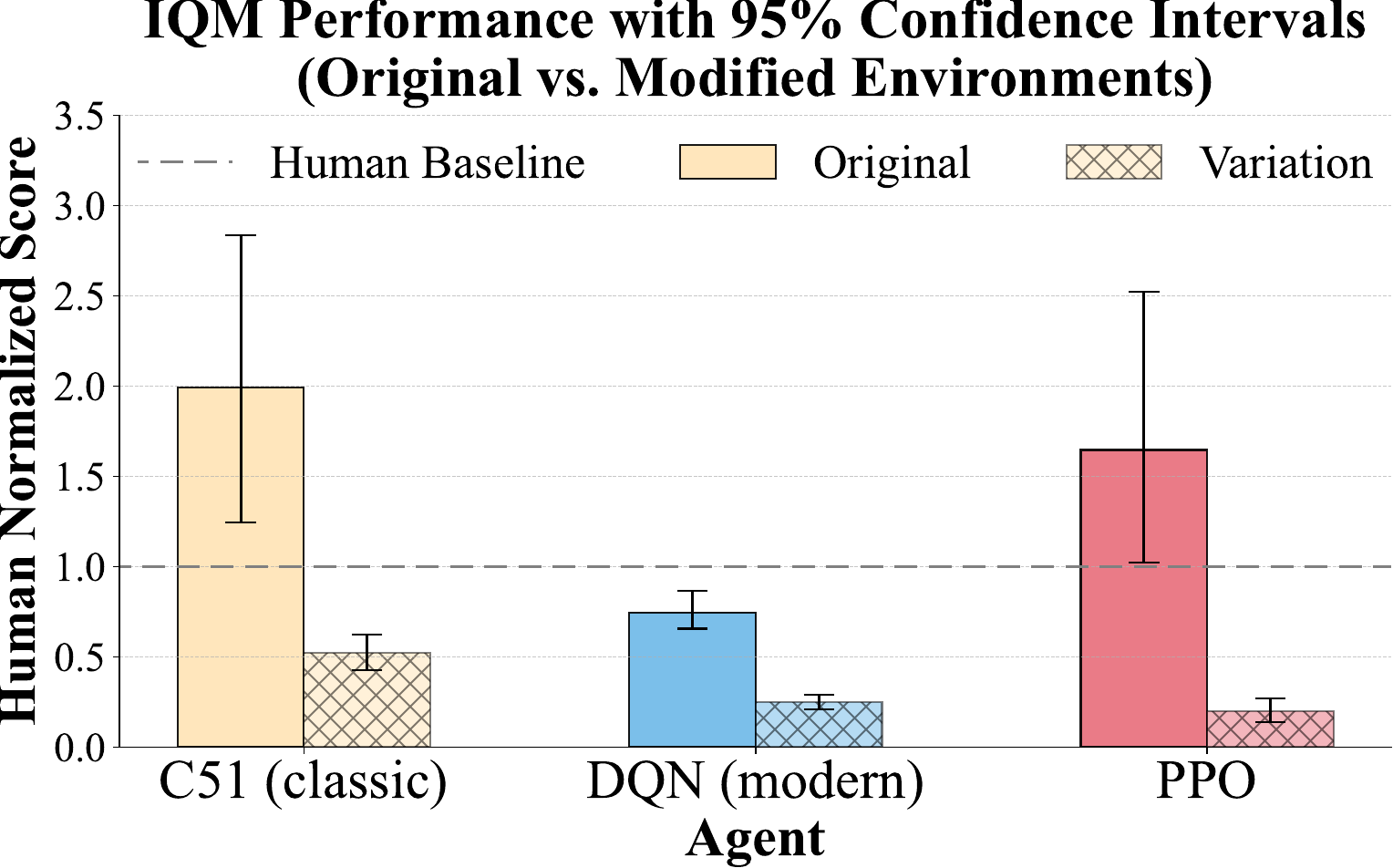}
    \caption{\textbf{Deep agents cannot generalize to simpler scenarios.} Testing deep agents in simpler or slightly changed versions of the environments reveals a significant performance drop, highlighting a critical limitation in their robustness and adaptability. The reported numbers are the interquartile mean (IQM) over 3 seeds, with 10 runs each on 11 games (C51 on six games). The exact performances are in \autoref{app:results}.}
    \label{fig:NotRobust}
    
\end{wrapfigure}

Reinforcement learning (RL) has predominantly relied on learning directly from raw pixel inputs without an explicit object extraction step, a paradigm introduced with Deep Q-Networks (DQN)~\citep{mnih2015human}. While this task-agnostic architecture has enabled various algorithmic advancements, end-to-end convolutional neural network (CNN)-based approaches often exhibit sensitivity to noise and spurious correlations, leading to policies that fail under distribution shifts \citep{agnew2021relevanceguidedmodelingobjectdynamics, yoon2023investigationpretrainingobjectcentricrepresentations, Hermann24Shortcut}. \citet{farebrother2020generalizationregularizationdqn} highlight that such agents frequently overfit to their training environments, struggling to generalize to task variations.
A particularly striking example is the failure of RL agents in Pong, where they learn to base their actions primarily on the opponent’s paddle position—a spurious shortcut—while largely ignoring the ball’s trajectory \citep{Hackatari, delfosse2024interpretable}. As illustrated in \autoref{fig:NotRobust}, CNN-based agents exhibit significant performance drops when evaluated outside their training environments. These limitations underscore the necessity of incorporating structured representations and inductive biases to improve generalization and robustness in RL.
Therefore, inspired by the results from cognitive science, recent studies have proposed integrating abstraction processes into RL to enhance generalization~\citep{zhao2021consciousness, bertoin2022look}. 
The object-centric approaches offer a promising solution by disentangling scenes into object-level attributes and relationships. 
Advances like SPACE~\citep{Lin2020Space} and Slot Attention~\citep{locatello2020slots}, notably applied to Atari RL environments~\citep{Delfosse2021MOC} demonstrate the potential of unsupervised object extraction methods that can extract object representations to improve efficiency and generalization~\citep{patil2024contrastiveabstractionreinforcementlearning}. 
Thus, many symbolic policies have been developed to take advantage of object-centric extraction ~\citep{delfosse2024interpretable, kohler2024interpretable, luo2024end, marton2024sympol}, with novel end-to-end object-centric neurosymbolic approaches that use limited~\citep{luo2024end} or even no supervision~\citep{grandien2024interpretable}.

However, integrating object extractors into RL agents presents several challenges. These methods often require expert-defined priors, such as the maximum number of objects, object categories, and relevant attributes (e.g., position, size, color, orientation, shape). Extracting excessive information introduces noise, increasing the risk of shortcut learning~\citep{kohler2024interpretable}, while excessive compression may discard critical task-relevant details, hindering learning a suitable policy. 
Moreover, object extraction models must be trained separately before RL can begin, adding computational overhead and limiting adaptability.

We introduce \textbf{O}bject-\textbf{C}entri\textbf{c} \textbf{A}ttention via \textbf{M}asking (OCCAM), an object-oriented abstraction method inspired by the principle of Occam’s Razor. OCCAM implements an object-centric attention bias by masking out non-object pixels from the input frames provided to CNN-based policies. This method relies on a simple object detection technique that only needs to extract object bounding boxes, avoiding complex pre-processing or fine-grained object analysis. By removing extraneous background information, OCCAM significantly reduces perceptual noise, limits reliance on spurious correlations, and enhances generalization across diverse environments.

Our contributions can be summarized as follows:
\begin{enumerate}
\item We propose Object-Centric Attention via Masking (OCCAM), a novel, task-agnostic object-oriented method inspired by Occam's Razor principle.
\item We provide a detailed discussion on the strengths and limitations of various representations utilized by RL agents in Atari, emphasizing the trade-offs involved in abstraction.
\item We empirically demonstrate that OCCAM achieves high performance, 
and significantly increases robustness against visual perturbations.
\item We question the necessity of symbolic extraction for addressing the Pong misalignment problem~\citep{delfosse2024interpretable}, showing that OCCAM can mitigate shortcut learning without requiring complete symbolic representations.
\end{enumerate}


\section{Rethinking Input Representations}

This section introduces our Object-Centric Attention via Masking (OCCAM) principle, a novel abstraction mechanism designed to enhance robustness and generalization. 
OCCAM's goal is to reduce the extreme dependency of classic CNN-based agents of irrelevant visual artifacts~\citep{ma2020discriminative} while avoiding the necessity of building complex tasks-specific symbolic systems that rely on expert knowledge specific to each environment.

\subsection{Object-Centric Attention via Masking}

\begin{figure}[t]
    \centering
    \includegraphics[width=\linewidth]{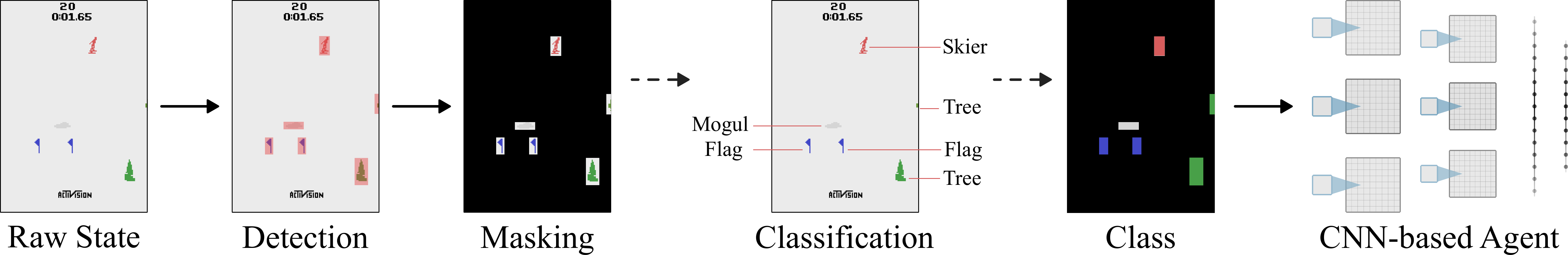} 
    \caption{\textbf{Object-Centric Attention via Masking.} An object detection method detects moving objects, allowing it to mask out the irrelevant background details. Optionally, objects can be classified to obtain a class-augmented mask. This representation is then passed to the CNN-based agent.}
    \label{fig:object_centric_attention}
\end{figure}

Inspired by Occam’s Razor—which advocates for minimal yet effective explanations—OCCAM enhances RL by constructing structured visual representations. OCCAM systematically reduces visual complexity by eliminating background clutter while preserving object information, all without requiring domain-specific priors or external annotations.
%
%
By applying a lightweight masking strategy (see \autoref{fig:object_centric_attention}), OCCAM refines input representations, reducing perceptual noise and minimizing potential confounders. This ensures that CNN-based deep RL architectures (e.g., DQN, PPO) process only the most relevant features.

Object extraction is crucial in OCCAM’s masking approach. Existing methods differ in their assumptions and design: motion-based approaches (e.g., Optical Flow~\citep{farneback2003two}) track dynamic elements but may struggle with static objects, while self-supervised feature learning (e.g., DINO~\citep{zhang2022dino}), YOLO~\citep{redmon2016you} and Mask R-CNN~\citep{he2017mask} groups objects based on learned representations. While these methods effectively extract object-like entities, symbolic extraction—which assigns objects high-level roles, relationships, or causal dependencies \citep{lake2017building, Bengio17Prior}—introduces an additional layer of abstraction that typically relies on structured priors. While this is needed for symbolic approaches, it also adds another layer of complexity that can introduce biases, limit adaptability, and reduce scalability.

In this work, we evaluate different masking strategies that can either rely on task-agnostic or task-specific representations, adapting to the available information. While fully task-agnostic masking enables generalization without predefined structures, when additional information—such as object categories or domain-specific knowledge—is accessible, OCCAM can integrate these details. This flexibility allows for scalable, adaptable, and efficient representations that improve RL generalization while maintaining computational efficiency.

A key advantage of OCCAM is its ease of integration, making it a practical alternative to conventional object-centric methods. Unlike symbolic approaches that rely on task-specific object extractors, OCCAM does not require pretraining or domain-specific priors.
\citet{delfosse2024interpretable}, however, argue that the color (of the ghosts) is a crucial feature to solve \textit{Pacman} and \textit{MsPacman} and that the orientation is necessary to solve \textit{Skiing}. Thus, their symbolic method relies on task-specific fine-tuned object extractors to extract such concepts.
In contrast, OCCAM allows the CNN to learn relevant object properties implicitly, reducing computational overhead while preserving a structured input. Moreover, most masking strategies maintain the same input dimensionality as DQN, ensuring compatibility with standard RL frameworks without requiring architectural modifications.

OCCAM is not limited to a single masking approach but presents a general framework for scene abstraction, emphasizing essential information while filtering irrelevant details. To evaluate this, we analyze a range of abstraction levels—from binary masks, which aggressively reduce visual complexity, to structured, task-specific masks that retain the object classes as properties. 

In summary, OCCAM provides robust and adaptable visual representations, balancing abstraction and expressivity to improve RL generalization in complex environments.

\subsection{OCCAMs Abstraction Levels}
\label{sec:masks}

\begin{figure}[tb]
    \centering
    \includegraphics[width=0.99\linewidth]{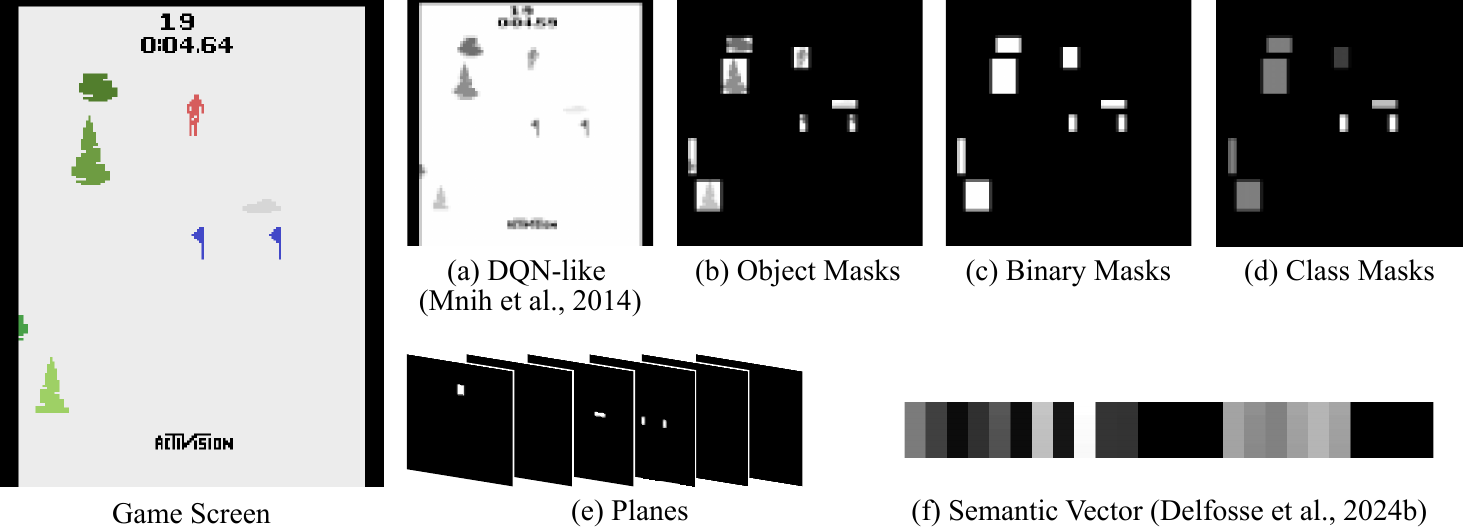}
    \caption{\textbf{The different extracted representations compared in this paper.} 
    For a given frame (left), (a) gray scaling and resizing are applied to reduce computational complexity. On top of it, (b) uses an object extractor to mask out the background information, (c) whitens the bounding boxes,
    (d) relies on a classifier to assign each object box a given class color, while (e) separates the masks in different planes for each class. 
    Agents are provided with stacks of the latest $4$ extracted representations, except for the semantic vector (f), which extracts symbolic properties of the depicted objects from the last $2$ frames. More examples are provided in \autoref{app:masks}.}
    \label{fig:masks}
\end{figure}

 Rather than presenting one new strategy, OCCAM supports multiple abstraction strategies that vary in visual detail and object-specific information. 
 Below, we present four masking approaches that are concrete implementations of OCCAM’s principles, as illustrated in \autoref{fig:object_centric_attention}. All masks are visualized in \autoref{fig:masks}.

\textbf{Object Masks:} Retains object positions and appearances while removing background information. This representation preserves all object-specific features, including color, ensuring minimal loss of information while filtering irrelevant elements.

\textbf{Binary Masks:} Represents objects in a binary occupancy grid, where pixels inside object bounding boxes are set to 1, and all other pixels are 0. This removes object information, reducing potential spurious correlations but eliminating fine-grained details that may be useful for decision-making.

\textbf{Class Masks:} Encodes objects by category while discarding all other visual attributes. Each object is assigned a uniform representation based on its class, preserving semantic structure but requiring an object extraction method capable of classification, making it task-dependent.

\textbf{Planes:} Inspired by structured representations in chess AI \citep{Browne14, silver2016mastering, czech24representation} and the work by \citet{Davidson20}, this format represents object categories as separate binary planes, where each plane contains only one object type. This structured encoding improves policy stability and generalization but increases input dimensionality, scaling with the number of object classes.

\section{Empirical Evaluation}

To assess the effectiveness and limitations of OCCAM, we design a structured empirical analysis addressing the following research questions:

\begin{enumerate}[label=\textbf{(Q\arabic*)}, leftmargin=*]


\item How does OCCAM perform compared to raw pixel-based and symbolic OC approaches?

\item Does OCCAM effectively reduce the tendency of RL agents to rely on irrelevant or spurious visual correlations?

\item Can we find the optimal balance between representational simplicity and task performance across different abstraction levels?
\end{enumerate}

We conduct experiments across multiple Atari environments to systematically evaluate OCCAM, focusing on three key aspects. First, we benchmark OCCAM’s object-centric representations against conventional pixel-based reinforcement learning using Proximal Policy Optimization (PPO)~\citep{schulman2017proximal} and OCAtari~\citep{OCAtari}, a symbolic object-centric baseline.
Second, we assess robustness to environmental variations by evaluating performance under standard and perturbed conditions. These perturbations, derived from HackAtari~\citep{Hackatari}, include visual modifications (e.g., color shifts, object recoloring) and gameplay variations (e.g., altered agent dynamics and enemy behavior).

A detailed description of the experimental setup, including environment selection, training procedures, and hyperparameter configurations, is provided in \autoref{sec:setup}.

\subsection{Improved Performance with less Input Information}
\begin{wrapfigure}[17]{r}{0.5\linewidth}
    \vspace{-5mm}
    \includegraphics[width=\linewidth]{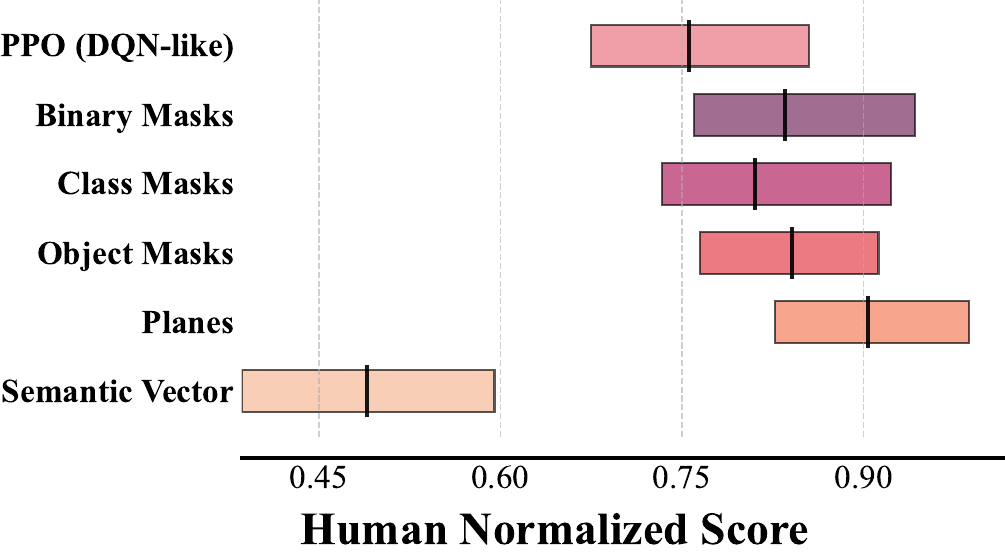}
    \caption{\textbf{OCCAM-based representations match or surpass pixel inputs, showing that abstraction improves performance.} This figure compares PPO agents using different input types. OCCAM preserves effectiveness despite filtering details, proving structured inputs can replace raw pixels. Full results are in \autoref{app:results}.}
    \label{fig:ppo_performance}
\end{wrapfigure}

To evaluate how object-centric attention impacts RL performance compared to raw visual inputs and object-centric ones, we analyze aggregated human normalized scores (HNS) across multiple Atari environments (more about the metrics in \autoref{app:metrics}. This addresses the first research question by assessing whether OCCAM-based representations are comparable or superior to existing baselines.
For this study, we compare six different input representations: the four masking approaches introduced in Section~\ref{sec:masks}, alongside two baselines:

\textbf{DQN-like}~\citep{MnihKSGAWR13}: A standard deep RL representation, particularly in Atari~\citep{fan21surveyatari}, where agents process downsampled grayscale pixel inputs in a stack of four. This representation is computationally expensive and relies on convolutional feature extraction. In this study, we use PPO~\citep{schulman2017proximal} as architecture in combination with this representation.

\textbf{Semantic Vector}~\citep{OCAtari}: A structured representation that encodes object attributes (e.g., position, size, and additional properties) for symbolic reasoning, independent of raw pixels. This format significantly reduces input dimensionality into a single vector.

The results in \autoref{fig:ppo_performance} show that OCCAM matches or outperforms our baselines across six standard environments.
These findings demonstrate that object-level information alone is sufficient for effective policy learning. Despite using a sparser input representation, OCCAM maintains or improves scores, indicating that removing task-irrelevant details does not degrade performance.

\subsection{Generalization under Environmental Variations: How OCCAM Handles Perturbations}  

\begin{figure}
    \centering
    \includegraphics[width=\linewidth]{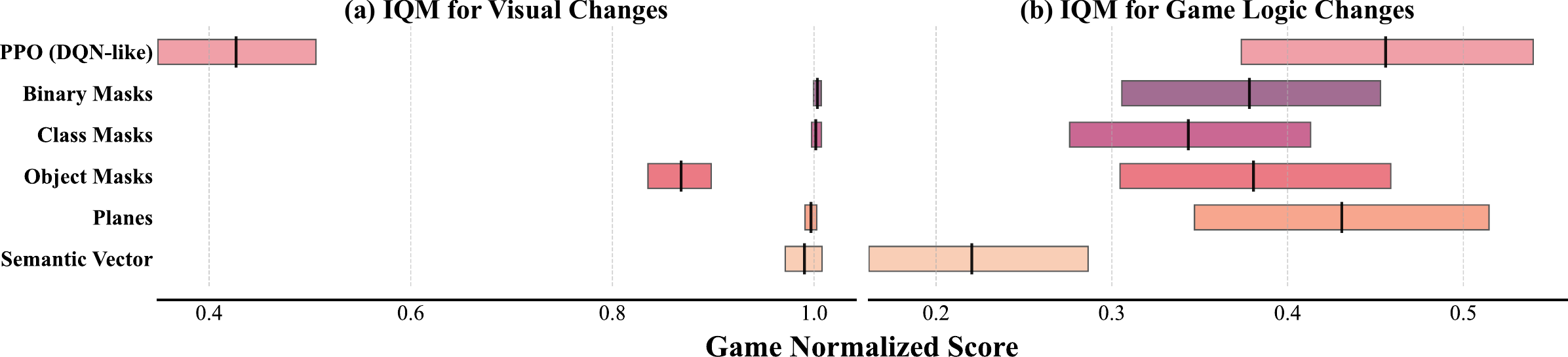}
    \caption{\textbf{OCCAM-based representations improve robustness to visual perturbations but remain vulnerable to game logic changes.} This figure illustrates the relative performance (game normalized score)  of PPO agents utilizing different input representations under both (a)~visual and (b)~game logic perturbations. While OCCAM-based representations significantly mitigate performance degradation due to visual modifications, they remain sensitive to changes in game mechanics, highlighting the limits of abstraction in RL. The game-specific scores are in \autoref{app:results}.}
    \label{fig:gns}
\end{figure}

While OCCAM’s object-centric representations improve robustness, they remain sensitive to certain environmental variations. To assess their generalization properties, we evaluate performance under two types of perturbations: visual modifications, which, e.g., alter object color while preserving game mechanics, and game logic alterations, which modify object behavior like movement patterns.
Rather than relying solely on the HNS, we measure relative performance changes between the original and perturbed environments---the game normalized score (GNS). This approach provides a more detailed understanding of how different representations adapt across varying conditions.

\textbf{Visual Robustness: OCCAM Representations Maintain Stability.}
As shown in \autoref{fig:gns}, OCCAM-based agents demonstrate robust performance under visual perturbations, validating the effectiveness of structured abstraction in RL. Representations such as Binary Masks and Planes, which remove background details while preserving object structures, allow agents to retain decision-making accuracy despite changes in object appearance. This result confirms that policies can generalize surface-level visual differences by focusing on underlying information such as position.

\textbf{Game Logic Sensitivity: Abstraction Alone Is Not Enough.}
While OCCAM-based representations reduce reliance on visual distractions, they remain vulnerable to changes in game mechanics. In Freeway, for example, alterations to vehicle behavior cause a significant drop in performance, indicating that learned policies depend on predictable motion patterns.
In Pong, structured representations—particularly the Planes—partially mitigate shortcut learning by enforcing object-centric reasoning rather than reliance on spurious correlations (cf. \autoref{tab:pong}). It illustrates that the agents that use DQN-like or semantic vector as input, which are trained on \textit{Pong}, suffer from immense performance degradations when evaluated in environments where the enemy stops moving when the ball is going to the agent (Lazy Enemy) or when the Enemy is hidden. \citet{Hackatari} have shown that deep agents obtain the same performances when trained on each environment variation. The different OCCAM variations suffer from much lower performance degradations.
Notably, OCCAM addresses this issue without requiring symbolic reasoning, contradicting the claim by \citet{delfosse2024interpretable} that symbolic extraction is necessary to resolve the Pong misalignment problem, suggesting that incorporating an object-centric and a locality bias reduces such shortcut dependence. 
Our conclusion can be seen from the perspective of~\citet{Hermann24Shortcut}, which explains that statistical (visual) agents learn to rely on the simplest decodable feature, not the most discriminative one. Here, the object-centric masking and locality importance prior (from the CNN) makes the agent rely on the ball's position to place its paddle instead of the correlated enemy's. As the ball is the object bouncing on the paddle, incorporating such a prior makes the agent robust to the decorrelated Lazy and Hidden Enemy variations. 
Nonetheless, this effect is inconsistent over multiple environments (cf. \autoref{fig:gns}). Performances can still be degraded for many types of task variations, and further work shall address these generalization problems.

\begin{table}[t]
    \centering
    \small
    \setlength{\tabcolsep}{3pt}  
    \renewcommand{\arraystretch}{1.1}  
    \caption{\textbf{Having the correct inductive bias can reduce misalignment.} Performance comparison of different representations in the Pong environment with and without a lazy enemy. The drop in performance (in percentage) highlights the impact of different input representations on robustness. The \tcbox[colback=red!30]{color gradient} emphasizes the severity of the drop, with stronger red indicating a greater decline in performance.  
    }
    \label{tab:pong}
    \resizebox{\linewidth}{!}{
\begin{tabular}{lccccccc}
\toprule
    Game & DQN-like & Object Masks & Binary Masks & Class Masks & Planes & Semantic Vector\\
\midrule
\underline{Pong} & $21.00 \ci{21, 21}$ & $21.00 \ci{21, 21}$ & $21.00 \ci{21, 21}$ & $21.00 \ci{21, 21}$ & $21.00 \ci{21, 21}$ & $21.00 \ci{21, 21}$ \\
\rowcolor[gray]{0.95}Lazy Enemy & \tcbox[colback=red!50]{$-4.38 \ci{-11, -0}$} & \tcbox[colback=red!5]{$20.00 \ci{20, 20}$} & \tcbox[colback=red!30]{$11.25 \ci{-2, 20}$} & \tcbox[colback=red!40]{$8.25 \ci{3, 14}$} & \tcbox[colback=red!30]{$12.44 \ci{-1, 20}$} & \tcbox[colback=red!60]{$-21.00 \ci{-21, -21}$} \\
\rowcolor[gray]{0.95}Hidden Enemy & {--} & { $21.00\ci{21, 21}$} & { $21.00\ci{21, 21}$} & { $21.00\ci{21, 21}$} & { $21.00\ci{21, 21}$} & \tcbox[colback=red!60]{$-20.88\ci{-21, -21}$} \\
\bottomrule
\end{tabular}
}

\end{table}

\subsection{The Trade-Off Between Abstraction and Expressivity in Visual Reasoning}
\begin{table}[t]
    \centering
    \small
    \setlength{\tabcolsep}{3pt}  
    \renewcommand{\arraystretch}{1.1}  
    \caption{\textbf{Removing the maze structure (MsPacman), river (Riverraid), or skier orientation (Skiing) does not degrade performance.} Despite reducing structural information, agents trained with Object and Binary Masks perform comparably, suggesting that task-relevant features remain sufficient for effective learning. \tcbox[colback=LightBlue]{Blue} marks all values within 5\% of the best performance. \\ \small $^\ast$Skiing has an ill-defined reward function, which needed adaptation in training (cf. Section \ref{app:skiing}).}
    \label{tab:toomuch}
    \resizebox{\linewidth}{!}{
    \begin{tabular}{lccccccc}
\toprule
    Game & DQN-like & Object Masks & Binary Masks & Class Masks & Planes \\
\midrule
\underline{MsPacman} & $3174.38 \ci{2792, 3637}$ & $5880.00 \ci{5531, 6068}$ & $4833.12 \ci{4583, 5221}$ & $4549.38 \ci{3810, 5188}$ & \tcbox[colback=LightBlue]{$7187.50 \ci{6779, 7415}$}\\  
\underline{Skiing}$^\ast$ & \tcbox[colback=LightBlue]{$-3203.62\ci{-3233, -3175}$} & \tcbox[colback=LightBlue]{$-3235.25\ci{-3261, -3206}$} & \tcbox[colback=LightBlue]{$-3263.50\ci{-3323, -3229}$} & \tcbox[colback=LightBlue]{$-3201.25\ci{-3229, -3176}$} & \tcbox[colback=LightBlue]{$-3206.31\ci{-3237, -3178}$}  
\\\midrule
\underline{Riverraid} & \tcbox[colback=LightBlue]{$7849.38 \ci{7434, 8181}$} & 
\tcbox[colback=LightBlue]{
$7938.75 \ci{7780, 8068}$} &
\tcbox[colback=LightBlue]{$8076.88 \ci{7872, 8351}$} & \tcbox[colback=LightBlue]{$8160.00 \ci{7998, 8274}$} & \tcbox[colback=LightBlue]{$7953.75 \ci{7841, 8188}$}\\
\rowcolor[gray]{0.95}Linear River & \tcbox[colback=LightBlue]{$6689.38 \ci{5086, 7879}$} & $2903.75 \ci{2750, 3179}$ & $2705.00 \ci{2690, 2724}$ & $2832.50 \ci{2680, 3080}$ & $2931.88 \ci{2761, 3172}$ \\
\bottomrule
\end{tabular}
}
\end{table}

A key challenge in designing effective RL representations is balancing abstraction and expressivity. While abstract representations enhance generalization and reduce reliance on spurious correlations, excessive simplification may remove critical task dynamics, limiting an agent’s ability to learn optimal policies. Conversely, overly detailed representations risk encoding spurious patterns, leading to shortcut learning and overfitting. 
To examine this trade-off, we analyze the results in \autoref{tab:toomuch}, which compare different visual abstraction strategies in MsPacman, Riverraid, and Skiing. These environments were chosen due to their differing spatial dependencies—MsPacman requires maze navigation, whereas Skiing relies on continuous movement based on the orientation of one player.

The results show that removing structural details—such as the maze layout in MsPacman, the river in Riverraid, or skier orientation in Skiing—does not significantly degrade performance. Agents trained with object masks outperform those using DQN-like representations in MsPacman, suggesting that explicit structural encoding is not always necessary for effective learning. Similar can be seen in Riverraid, where the river can be learned by playing the game. This, however, means changing the river layout can drastically impact the performance (cf. \autoref{tab:toomuch}.  
In Skiing, performance remains unchanged mainly across representations, indicating again that explicit orientation encoding has little impact when task-relevant object information is retained, or information can be deduced implicitly using the temporal component in the stack of images provided. 

The performance drop in the Semantic Vector representation highlights the limitations of excessive compression, as it removes spatial dependencies, impairing the agent’s ability to model object relationships. This confirms that while symbolic representations improve interpretability, they struggle in tasks requiring fine-grained spatial or visual reasoning tasks.

Overall, these results indicate that abstraction remains effective as long as essential task dynamics are preserved. The optimal level of abstraction depends on the environment’s structure and the agent’s capacity to infer missing information from temporal or relational cues.

\clearpage
\paragraph{Choosing the Right Masking Strategy.}
There is no universally optimal masking strategy—each has trade-offs depending on the environment and task requirements. Planes generally perform best, providing structured spatial representations that improve generalization. However, they increase input size and computational cost, which may not be ideal for efficiency-constrained applications. Among the other masking approaches, Class Masks require additional pre-processing to categorize objects but do not consistently offer a clear advantage over simpler alternatives. Object Masks retain more visual details, benefiting environments where fine-grained features matter. However, it also tends to be easier to find misaligned due to the additional potential confounders. Binary Masks enforce more substantial abstraction, reducing reliance on texture and color but potentially discarding helpful information, like the color of the ghosts in MsPacman. Ultimately, the choice depends on the balance between abstraction and expressivity and the specific demands of the task at hand. Also, this paper is not about the concrete mask and more about the idea.

\section{Related Work}
\label{sec:related}

\textbf{Cognitive Science and Abstraction in RL}
Human cognition relies on abstraction, selectively filtering irrelevant details while retaining essential task-relevant information for decision-making~\citep{Bengio17Prior, goyal2022inductive}. In RL, \citet{zhao2021consciousness} demonstrated that focusing on key features enhances generalization across varying environments. Building on this, \citet{alver2024an} introduced an attention-based RL framework that prioritizes relevant visual elements. Our work aligns with these efforts by proposing a structured abstraction technique that enables RL agents to filter unnecessary information while efficiently preserving critical features for decision-making.

\textbf{State Representation Learning in RL.}
Effective state representations are crucial for RL agents operating in high-dimensional environments. Conventional methods, such as those in DQN~\citep{MnihKSGAWR13} and AlphaGo~\citep{Schrittwieser2019MasteringAG}, encode raw images into fixed-dimensional feature vectors. More structured approaches include feature-plane representations and scene graphs for modeling object relationships~\citep{YangLLBP18, KonerLHDTG21}. While these methods offer advantages, they often incur high computational costs and require task-specific tuning \citep{locatello2020slots, shengyi2022the37implementation, FarebrotherOVTC24}, limiting their scalability.

\textbf{Masking and Attention-Based Representations in RL.}
Selective attention mechanisms have been explored to enhance RL efficiency by focusing on task-relevant information~\citep{MottZCWR19, ManchinAH19}. Prior work has used attention-driven feature selection~\citep{XuBKCCSZB15}, spatial transformers~\citep{JaderbergSZK15}, and dynamic masking strategies~\citep{ManchinAH19} to refine input representations. While these methods rely on learned attention weights, OCCAM provides a structurally defined abstraction that could be described as hard attention. Unlike learned attention, OCCAM applies a predefined masking strategy that explicitly filters task-irrelevant details without relying on gradient-based optimization. This structured approach avoids instability and computational overhead caused by attention-based feature selection.

\textbf{Object-Centric RL.}
Object-centric RL focusing on objects and their interactions rather than raw pixels \citep{OCAtari, patil2024contrastiveabstractionreinforcementlearning}. Techniques such as Slot Attention~\citep{locatello2020slots} enable agents to decompose visual scenes into object-level representations, improving generalization and interpretability. Prior work has explored object-centric representations for RL generalization~\citep{DittadiPVSWL22, yoon2023investigationpretrainingobjectcentricrepresentations} and applied them to Atari environments~\citep{Hackatari, OCAtari, Davidson20}. However, existing approaches often suffer from excessive compression, which can obscure spatial relationships crucial for decision-making. OCCAM addresses this trade-off by selectively preserving object structure while reducing unnecessary visual details, ensuring that abstraction does not come at the cost of expressivity.

\textbf{Robustness and Generalization in RL.}
Ensuring robustness to environmental perturbations and domain shifts remains a key challenge in RL. Deep RL agents trained on raw pixels often overfit to spurious correlations, leading to catastrophic failures under minor changes in color or object placement, particularly in Atari benchmarks~\citep{farebrother2020generalizationregularizationdqn, Hackatari}.

Existing generalization techniques, such as data augmentation~\citep{yarats2021image}, domain randomization~\citep{tobin2017domain}, and invariant feature learning~\citep{zhang2021learning}, require extensive tuning and struggle with out-of-distribution scenarios. In contrast, object-centric representations improve robustness by providing structured inputs that remain stable across visual variations.
Our work extends these insights by integrating object-centric world models with adaptive filtering, ensuring RL agents focus on task-relevant elements and enhance robustness and interpretability.

\section{Conclusion}

This work examines how structured abstraction can improve generalization, robustness, and efficiency in reinforcement learning by reducing reliance on spurious correlations. Standard RL methods that process raw pixel inputs often capture irrelevant details, leading to overfitting and shortcut learning. In contrast, OCCAM applies object-centric attention, filtering out background noise while preserving task-relevant objects, allowing agents to focus on decision-critical elements.
As shown in \autoref{fig:ppo_performance} and \ref{fig:gns}, OCCAM-based representations match or exceed pixel-based approaches, particularly in robustness to visual perturbations. By enforcing structured abstraction, OCCAM enhances policy learning, demonstrating that removing task-irrelevant information does not degrade performance.

\textbf{Limitations and Future Directions.}
While OCCAM reduces visual distractions, it remains sensitive to game logic perturbations, indicating that abstraction alone is insufficient for robust generalization. The effectiveness of different abstraction levels varies across tasks, suggesting that optimal representation strategies depend on task demands. Additionally, symbolic representations, such as OCAtari’s vectorized encoding, struggle with spatial dependencies, leading to weaker performance in tasks requiring precise spatial reasoning. These findings highlight the trade-off between abstraction and expressivity, where preserving essential task dynamics is crucial for decision-making.

Future work should focus on adaptive abstraction mechanisms that dynamically adjust filtering based on task complexity and policy uncertainty. One approach is task-aware masking, which adapts to environmental conditions to improve generalization against game logic changes. This could involve temporal abstraction to track object interactions or causal reasoning to model dependencies. Further investigation into adaptive representation-policy learning could refine the balance between abstraction and expressivity, ensuring robust policy optimization across diverse environments.

This work contributes to developing more robust and generalizable RL agents that can operate effectively across varying environments by structuring input representations around task-relevant abstraction.

\subsubsection*{Broader Impact Statement}
\label{sec:broaderImpact}

Object-centric representations in reinforcement learning improve generalization and interpretability, with potential applications in robotics, autonomous systems, and high-stakes decision-making. However, their effectiveness depends on accurate object extraction, which may introduce biases if objects are misidentified or omitted. Additionally, in dynamic environments, reliance on object-centric features could lead to unexpected failures if critical task information is lost during abstraction.
While these risks can be mitigated through careful design and evaluation, further research is required to enhance robustness across diverse settings and prevent unintended consequences. Investigating adaptive filtering mechanisms and more flexible abstraction strategies could improve reliability in real-world deployments.
\vfill
\clearpage
\appendix

\section{Experimental Setup}
\label{sec:setup}
To evaluate the effectiveness of object-centric abstraction in RL, we conduct a series of controlled experiments using the Atari Learning Environment (ALE). We benchmark our approach against conventional pixel-based RL methods, including Deep Q-Networks (DQN)~\citep{MnihKSGAWR13}, Proximal Policy Optimization (PPO) \citep{schulman2017proximal}, and C51~\citep{bellemare2017distributional}, as well as OCAtari~\citep{OCAtari} as another object-centric representation with focus on symbolic representation.
The DQN and C51 baseline models were taken from \citet{gogianu2022agents}, while the PPO and OCAtari ones were trained ourselves.
To test robustness, we decided to test the trained agents not only in their original environment but also in their perturbations. These perturbations, derived from HackAtari~\citep{Hackatari}, include visual alterations (e.g., color changes, object recoloring), structural modifications (e.g., object displacement, swapped game elements), and gameplay variations (e.g., altered agent dynamics or enemy behavior). The metrics we were using are the human normalized score (HNS) and the game normalized score (GNS), both explained in \autoref{app:metrics}, over the aggregated game scores, using the interquartile mean (IQM). Calculations are done with rliable~\citep{agarwal2021rliable}.

\subsection{Environment Selection}
We evaluate our framework across a diverse set of 12 Atari games selected to balance reactive and strategic decision-making tasks (Boxing vs Freeway), having environments where the background information or object features are crucial for the game (MsPacman and Skiing) as well as some classics, such as Pong or Breakout. This selection allows us to assess the adaptability of different representations across a spectrum of task complexities. To list all environments again, we used Amidar, BankHeist, Bowling, Boxing, Breakout, Freeway, Frostbite, MsPacman, Pong, Riverraid, Skiing, and SpaceInvaders.

\subsection{Training PPO}
PPO is a policy gradient algorithm widely used in deep RL that optimizes a clipped surrogate objective to balance exploration and stability \citep{schulman2017proximal}. Unlike value-based approaches such as DQN, PPO directly learns an optimal policy distribution and is known for its sample efficiency and stable convergence properties. It is one of the most common architectures used in RL and, as such, is a good baseline as well as a base for our experimental section. 

We trained six PPO agents for the experiments using varying input representations. This allowed us to isolate the trade-offs between abstraction strength and spatial reasoning capacity and evaluate our visual reasoning with object-centric attention.
All agents are trained on the unmodified versions of these environments for 40 million frames using PPO, with hyperparameters adapted from \citet{huang2022cleanrl} and listed below. For PPO we adhere to standardized implementation guidelines \citep{shengyi2022the37implementation} to ensure comparability and used the aforementioned CleanRL framework~\citep{huang2022cleanrl} for training. The object-centric representations, derived using OCAtari, selectively preserve task-relevant features while eliminating extraneous background information. Importantly, no fine-tuning is performed in perturbed environments, ensuring that generalization performance reflects the robustness of the learned representations rather than additional adaptation. 

Performance is evaluated using average episodic rewards across 10 games per seed, with three fixed seeds per experiment. All experiments are conducted on NVIDIA A100 GPUs, with training times averaging around 2--4 hours per agent per seed for the DQN-like, Object Mask, Binary Mask, and Class Mask. Training the Semantic Vector representation took less than 2h (avg. 1h 38min), while for Planes, the environment highly influenced the training time and took 6--10h per game per seed.

\subsection{Hyperparameters}
We detail the hyperparameters used in our Proximal Policy Optimization (PPO) training to ensure reproducibility and consistency in our experiments. These settings were chosen based on prior literature.

\begin{table}[h]
\centering
\setlength{\tabcolsep}{3pt}  
\renewcommand{\arraystretch}{1.1}  
    
\caption{Key hyperparameters used for PPO training.}
\label{tab:hyperparams}
\begin{tabular}{|l|c||l|c|}
\hline
\textbf{Hyperparameter} & \textbf{Value} & \textbf{Hyperparameter} & \textbf{Value} \\
\hline
Seed & \{0,\,1,\,2\} & 
Learning Rate ($\alpha$) & $2.5 \times 10^{-4}$ \\
Total Timesteps & $10^7$ &
Number of Environments & $10$ \\
Batch Size ($B$) & $1280$ &
Minibatch Size ($b$) & $320$ \\
Update Epochs & $4$ &
GAE Lambda ($\lambda$) & $0.95$ \\
Discount Factor ($\gamma$) & $0.99$ &
Value Function Coefficient ($c_v$) & $0.5$ \\
Entropy Coefficient ($c_e$) & $0.01$ &
Clipping Coefficient ($\epsilon$) & $0.1$ \\
Clip Value Loss & \texttt{True} & 
Max Gradient Norm ($\|g\|_{\text{max}}$) & $0.5$ \\
\hline
\end{tabular}
\end{table}

\begin{table}[h]
\centering
\caption{Key hyperparameters regarding the used environments. We are using Gymnasium and the ALE, following the best practices by \citet{MachadoBTVHB18}.}
\setlength{\tabcolsep}{3pt}  
\renewcommand{\arraystretch}{1.1}  
    
\begin{tabular}{|l|c||l|c|}
\hline
\textbf{Hyperparameter} & \textbf{Value} & \textbf{Hyperparameter} & \textbf{Value} \\
\hline
ALE version & 0.8.1 & 
Gymnasium version & 0.29.1 \\
Environment version & v5 &
Frameskip & 4 \\
Buffer Window Size& 4 (PPO), 2 (Sem. Vector) &
Observation Mode & RGB \\
Repeat Action Probability & 0.25 (Training), 0 (Testing) &
Full Action Space & \texttt{False} \\
Continuous & \texttt{False} & & \\
\hline
\end{tabular}
\label{tab:hyperparams2}
\end{table}

\section{Reproducibility Statement}
We provide full experimental details to facilitate reproducibility, including hyperparameter configurations, random seeds, and training scripts. Each model is trained with three independent seeds (0, 1, 2) to ensure statistical robustness and account for variance in reinforcement learning training. Our implementation follows the CleanRL framework \citep{huang2022cleanrl}, a well-established reinforcement learning library designed for transparency, simplicity, and ease of replication.

\paragraph{Code and Data.}
Our masking approaches are implemented as wrappers for the OCAtari/HackAtari environments \citep{Hackatari, OCAtari}, as they provide a consistent object extraction that is easy to use for Atari games as well as an easy way to adapt and create perturbations. Our code and some models to test it can be found under \url{https://github.com/VanillaWhey/OCAtariWrappers}.

\paragraph{Test it yourself.}
To test our approach, we provide a small set of models that can be used with the provided run, print, and evaluation scripts (see scripts folder in the repository) to visualize the results. With the scripts, you can measure the performance reported in the paper and test other perturbations and games. Our training is based on a slight adaptation of the CleanRL framework. 

To run the evaluation script with the correct perturbations, see the commands in \autoref{app:variants}.

\clearpage

\subsubsection*{Acknowledgments}
\label{sec:ack}
We would like to express our gratitude for funding our project to the German Federal Ministry of Education and Research and the Hessian Ministry of Higher
Education, Research, Science and the Arts (HMWK). This project was enabled due to their joint support of the National Research Center
for Applied Cybersecurity ATHENE, via the ``SenPai: XReLeaS'' project and the cluster project within the Hessian Center for AI (hessian.AI) ``The Third Wave of Artificial Intelligence - 3AI''. 



\bibliography{bib}
\bibliographystyle{rlj}

\beginSupplementaryMaterials

\section{Human Normalized and Game Normalized Scores}
\label{app:metrics}

In this work, we used two widely used metrics comparing the performance of agents in Atari: the IQM over human normalized scores (HNSs) and the game normalized score (GNS) over the game scores, aggregated with IQM, using rliable \citep{agarwal2021rliable}. Both provide standardized performance evaluations across different environments.

\subsection{Human Normalized Score (HNS)}

The HNS is a metric that compares an RL agent's performance to that of a human expert and a random agent. It is computed as follows:
\begin{equation}
    \text{HNS} = \frac{\text{Score}_{\text{agent}} - \text{Score}_{\text{random}}}{\text{Score}_{\text{human}} - \text{Score}_{\text{random}}}
\end{equation}
where:
\begin{itemize}
    \item $\text{Score}_{\text{agent}}$ is the episodic score achieved by the RL agent.
    \item $\text{Score}_{\text{random}}$ is the score obtained by a random policy.
    \item $\text{Score}_{\text{human}}$ is the score achieved by a human expert.
\end{itemize}

An HNS value of $1.0$ means the agent matches human performance. Values close to $0$ suggest the agent performs at a level similar to random play. In this work, we assumed that our perturbations only slightly change the difficulty and performance of the human expert. Hence, we assumed the same human score for all perturbations in the original game. We agree that evaluating humans on perturbations would be the correct way, but it requires that the possibility exists. 

\subsection{Game Normalized Score (GNS)}

Unlike HNS, which compares performance to a human baseline, the GNS measures an agent’s performance relative to its performance in the original (unperturbed) environment. This ensures that an agent playing in the original environment always receives a score of $1.0$, while agents evaluated in modified environments are measured relative to this baseline:
\begin{equation}
    \text{GNS} = \frac{\text{Score}_{\text{agent}}}{\text{Score}_{\text{original}}}
\end{equation}
where:
\begin{itemize}
    \item $\text{Score}_{\text{agent}}$ is the episodic score achieved by the RL agent in the modified environment.
    \item $\text{Score}_{\text{original}}$ is the agent’s performance in the original, unperturbed environment.
\end{itemize}

With a GNS of $1.0$, the agent performs identically to its original environment performance. Scores above $1.0$ indicate improved performance in the new environment, while scores below $1.0$ signify a performance drop.

\subsection{Interpretation and Application}

\begin{itemize}
    \item The HNS metric provides a way to evaluate whether an RL agent is reaching or surpassing human-level play. It is widely accepted and used in a broad range of works, making the results of this work comparable. 
    \item The GNS metric ensures a fair comparison of robustness across different environmental conditions by using the original performance as a reference.
\end{itemize}

\section{Variants Overview}
\label{app:variants}

\begin{figure}[h]
    \centering
    \includegraphics[width=0.8\linewidth]{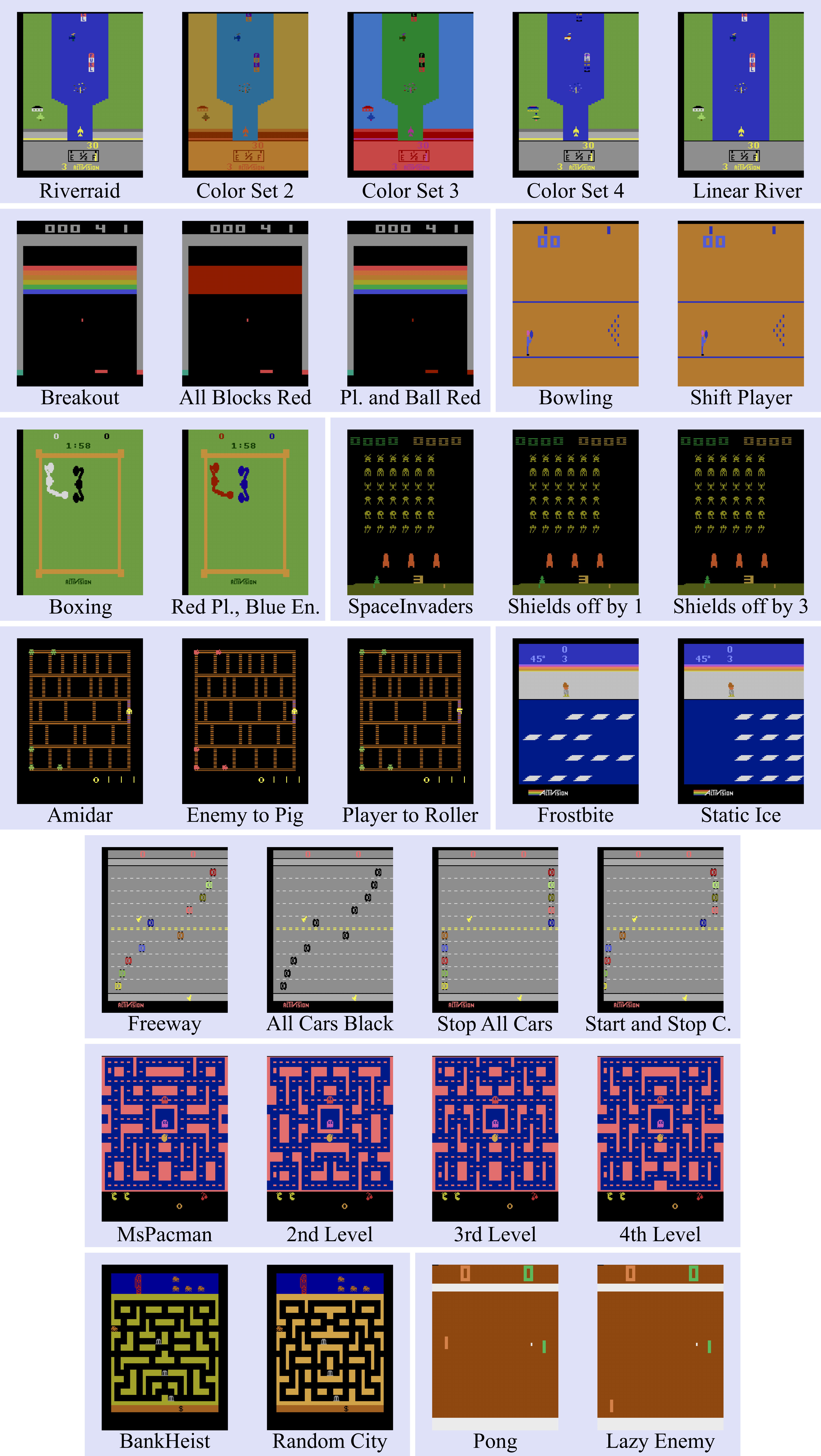}
    \caption{Illustration of different game variants used in our study. Games are grouped on a blue-shaded background, starting with the original game on the left.
    These variants introduce modifications such as color changes and (opponent) behavior variations to study their impact on learning and generalization.}
    \label{fig:variants}
\end{figure}

Below, we briefly describe each variant used in our study. Descriptions are done by us or taken from the ALE Documentation~\citep{GymAtari}. These variants are visualized in \autoref{fig:variants} and created using the HackAtari Environment~\citep{Hackatari}. The command needed to start the eval or run script with this variant is given below. This can be used to visualize or evaluate the models' performances.
\begin{minted}[fontsize=\small, tabsize=1]{bash}
python scripts/eval.py -g $GAME -a $MODEL_PATH -m $MODIFICATION_LIST
\end{minted}
,e.g.,
\begin{minted}[fontsize=\small, tabsize=1]{bash}
python scripts/eval.py -g Boxing -a models/Boxing/0/model.py \
-m color_player_red color_enemy_blue
\end{minted}

\subsection{Amidar Variants}
\textbf{Amidar:} \textit{Amidar} is similar to Pac-Man: You are trying to visit all places on a 2-dimensional grid while simultaneously avoiding your enemies.

\textbf{Enemy to Pig:} Change the enemy warriors into Pigs. The game logic stays the same. 

\textbf{Player to Roller:} Changing the sprite of the player figure from a human to a paint roller. The game logic stays the same.

\subsection{BankHeist Variants}
\textbf{BankHeist:} You are a bank robber and (naturally) want to rob as many banks as possible. You control your getaway car and must navigate maze-like cities. The police chase you and will appear whenever you rob a bank. You may destroy police cars by dropping sticks of dynamite. You can fill up your gas tank by entering a new city. This is your \textit{BankHeist}.

\textbf{Random City:} The starting city and the city you enter are randomly selected and do not follow a pattern.

\subsection{Bowling Variants}
\textbf{Bowling:} Your goal is to score as many points as possible in the game of \textit{Bowling}. A game consists of 10 frames, and you have two tries per frame. 

\textbf{Shift Player:} The player starts closer to the pins.

\subsection{Boxing Variants}
\textbf{Boxing:} The standard \textit{Boxing} environment where two players compete to land punches.

\textbf{Red Player, Blue Enemy:} A modified version where one player is red and the other is blue, potentially influencing object perception.

\subsection{Breakout Variants}
\textbf{Breakout:} The original \textit{Breakout} environment where the player controls a paddle to break bricks. 

\textbf{All Blocks Red:} All blocks are red, removing the color of blocks which does not hold game-relevant information.

\textbf{Player and Ball Red:} The paddle is a little redder than in normal, potentially changing agent perception and behavior.

\subsection{Freeway Variants}
\textbf{Freeway:} The standard \textit{Freeway} environment where a chicken crosses a highway with moving cars. 

\textbf{All Black Cars:} All vehicles are black, reducing visual diversity between the cars. There are no additional changes.

\textbf{Stop All Cars:} All cars are stopped on the edge of the frame, making it trivial to pass the street.


\subsection{Frostbite Variant}
\textbf{Frostbite:} The original \textit{Frostbite} environment where the player builds an igloo while avoiding hazards. To collect ice, the player has to jump between moving ice shelves.

\textbf{Static Ice:} Ice platforms remain fixed instead of moving, altering difficulty (making it much easier). 

\subsection{MsPacman Variants}
\textbf{MsPacman} The standard \textit{MsPacman} environment that is very similar to the Pacman environment.

\textbf{Level X:} A later level with a changed maze structure. No agent ever reached this level in training. As such, it is out of distribution for all agents.

\subsection{Pong Variants}
\textbf{Pong:} The standard \textit{Pong} environment where two paddles hit a ball back and forth. 

\textbf{Lazy Enemy:} The opponent stays still while the ball is flying away from it and only starts moving after the player hits the ball. No visual changes are visible. This was presented as one of two examples of misalignment in \citet{delfosse2024interpretable}. 

\textbf{Hidden Enemy:} The opponent is hidden for the player. The only observable objects are the paddle and the Ball. This follows the experiment by \citet{delfosse2024interpretable} and is used to compare to their work directly. 

\subsection{Riverraid Variants}
\textbf{Riverraid:} In \textit{Riverraid}, you control a jet that flies over a river: you can move it sideways and fire missiles to destroy enemy objects. Each time an enemy object is destroyed, you score points (i.e., rewards).

\textbf{Color Set X:} Change the color of all objects to a different preset color—no change in the game logic.

\textbf{Linear River:} Change the river always to have the same shape, a single line, and always the same width—no splits, etc.

\subsection{SpaceInvaders Variants}
\textbf{SpaceInvaders:} In \textit{SpaceInvaders}, your objective is to destroy the space invaders by shooting your laser cannon at them before they reach the Earth. The game ends when all your lives are lost after taking enemy fire or when they reach the earth. 

\textbf{Shields off by X:} The shields are moved X pixel(s) to the right. No further changes.

\clearpage

\section{Representations in MsPacman and Frostbite }
\label{app:masks}

\autoref{fig:masks} displays the input representations used in our experimental section, illustrating their differences and highlighting which elements are retained or abstracted. This section presents two additional examples, applying the exact representations to different games to further examine their impact. By comparing these variations, we can better understand how various levels of abstraction influence learning across diverse environments.

\begin{figure}[h]
    \centering
    \includegraphics[width=\linewidth]{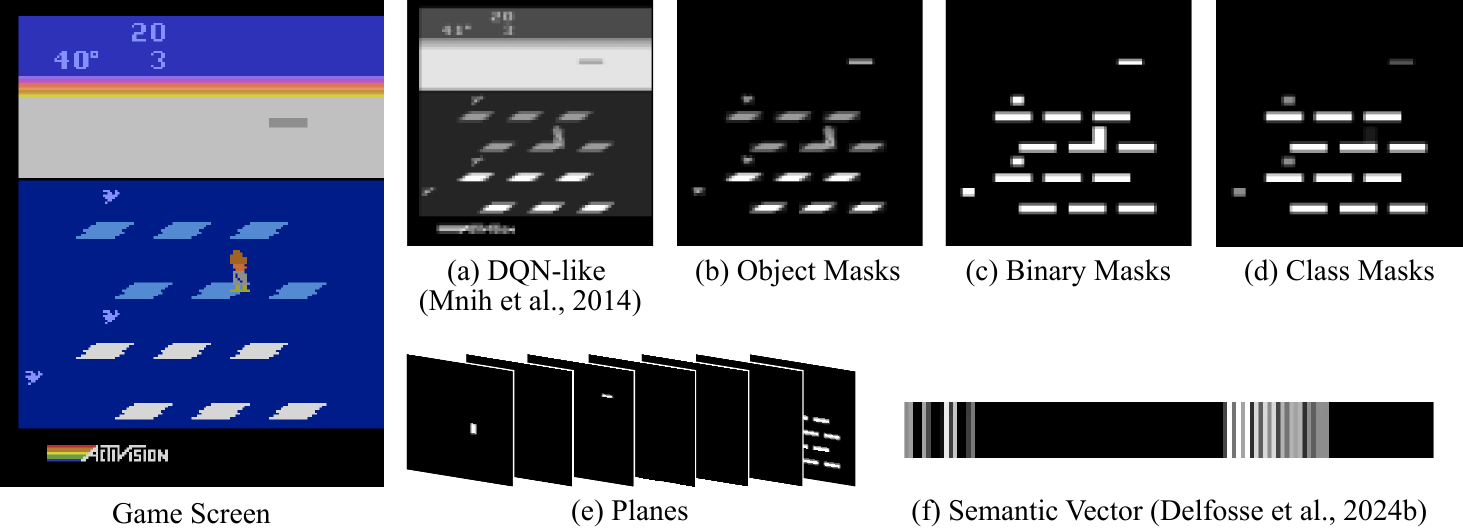}
    \caption{Alternative to \autoref{fig:masks}, showcasing the input representations using Frostbite as an example. This visualization highlights how different representations preserve or abstract various elements within the game environment}
    \label{fig:masks_frostbite}
\end{figure}

\begin{figure}[h]
    \centering
    \includegraphics[width=\linewidth]{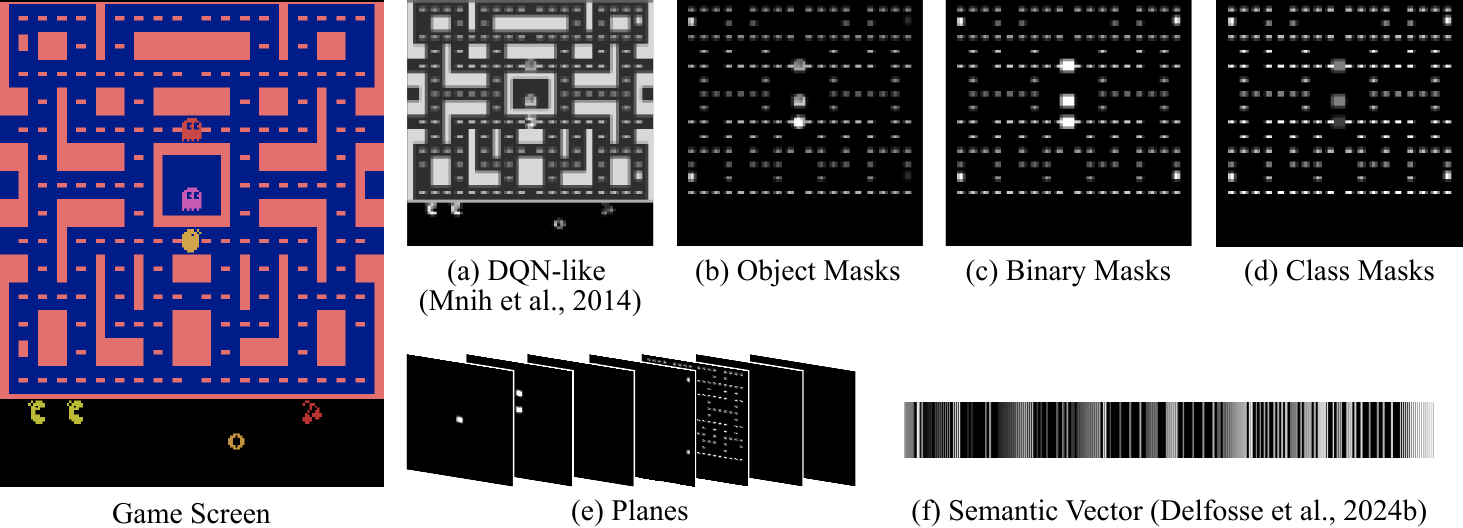}
    \caption{Alternative to \autoref{fig:masks}, showcasing the input representations using MsPacman as an example. This visualization highlights how the masking approach can visualize and take in small objects incorrectly. }
    \label{fig:masks_mspacman}
\end{figure}

\clearpage
\section{Additional/Extended Results}
\label{app:results}

\subsection*{Evaluation Setup}

We evaluate all agents using deterministic versions of the Arcade Learning Environment (ALE) to ensure a controlled comparison across input representations and perturbed variants. All agents are evaluated on three seeds with 10 episodes per seed per game. For the evaluation, we follow \citet{MachadoBTVHB18}. However, contrary to this, we disable sticky actions during evaluation by setting the repeat action probability to zero. While sticky actions typically introduce stochasticity for robustness assessments, our primary goal is to measure the agent’s ability to generalize across structured, deterministic modifications—such as visual or behavioral changes—without confounding effects from input-level randomness. As such, our evaluation setup is designed to isolate the effects of representation and abstraction. Results with sticky actions are still reported to enable comparison with other work and to check the robustness of the created policies against noise. Still, they are not further discussed in this work. 

\subsection*{Results}

\autoref{tab:results} presents the average performance of different RL agents—DQN, C51, and PPO—across a selection of Atari environments. The agents were trained exclusively in the standard environment (underlined) and evaluated on both their original settings and perturbed variations (gray-shaded). The perturbations include modifications such as color shifts, altered object properties, and behavioral changes designed to assess each method's robustness and generalization capabilities.

The results show that all agents perform relatively well in the standard environments, but experience varied performance degradation in perturbed settings. DQN, for instance, shows significant instability in perturbed settings, especially in environments like \textit{Red Player, Blue Enemy} (Boxing), and \textit{Player and Ball Red} (Breakout), where performance drops substantially. C51, while more stable in some cases, also struggles with drastic changes, particularly in \textit{Stopped Ice} (Frostbite). PPO is the most resilient, maintaining reasonable performance across some perturbed environments but still showing declines in all cases. These observations indicate that current pixel-based deep RL agents tend to overfit visual features during training, reducing adaptability when confronted with environment variations.

\begin{table}[h]
    \centering
    \small
    \setlength{\tabcolsep}{4pt}  
    \renewcommand{\arraystretch}{1.1}  
    
    \caption{
    This table presents the average performance (interquartile mean + 95\% confidence interval) of different RL agents evaluated in \textit{standard} (\underline{underlined}) and \textit{perturbed} (gray-shaded) environments. 
    All agents were trained only in the standard setting for 40M frames and evaluated on three seeds, playing 30 games per environment. A visualized version of this can be seen in \autoref{fig:NotRobust}.}
    
    \resizebox{.7\linewidth}{!}{
    \begin{tabular}{lllc}
\toprule
Game (Variant) & DQN & C51 & PPO \\ \midrule
\underline{Amidar} & $389.25\ci{258, 565}$ & -- & $564.75 \ci{540, 575}$ \\
\rowcolor[gray]{0.95} Enemy to Pig & $314.25\ci{207, 503}$ & -- & $555.00 \ci{543, 562}$ \\
\rowcolor[gray]{0.95} Player to Roller & $70.81\ci{58, 82}$ & -- & $149.75 \ci{124, 178}$ \\ \midrule
\underline{BankHeist} & $1281.25\ci{1215, 1326}$ & -- & $1176.88 \ci{1139, 1212}$ \\
\rowcolor[gray]{0.95} Random City & $1228.75\ci{1158, 1286}$ & -- & $1165.00 \ci{1131, 1198}$ \\ \midrule
\underline{Bowling} & $31.00\ci{24, 36}$ & $29.25\ci{23, 38}$ & $67.00 \ci{65, 70}$ \\
\rowcolor[gray]{0.95} Shift Player & $31.69\ci{23, 38}$ & $39.06\ci{34, 43}$ & $66.94 \ci{64, 70}$ \\ \midrule
\underline{Boxing} & $96.75\ci{94, 99}$ & $83.62\ci{80, 87}$ & $96.31 \ci{95, 98}$ \\
\rowcolor[gray]{0.95} Red Player, Blue Enemy & $-2.50\ci{-7, 1}$ & $-9.19\ci{-13, -5}$ & $-2.38 \ci{-10, 5}$ \\ \midrule
\underline{Breakout} & $177.25\ci{147, 208}$ & $378.75\ci{371, 385}$ & $290.25 \ci{259, 307}$ \\
\rowcolor[gray]{0.95} All Blocks Red & $277.44\ci{244, 308}$ & $344.50\ci{334, 358}$ & $306.31 \ci{277, 332}$ \\
\rowcolor[gray]{0.95} Player and Ball Red & $13.62\ci{11, 20}$ & $12.50\ci{10, 16}$ & $52.19 \ci{27, 89}$ \\ \midrule
\underline{Freeway} & $27.06\ci{16, 34}$ & $34.00\ci{34, 34}$ & $31.75 \ci{30, 32}$ \\
\rowcolor[gray]{0.95} All Cars Black & $12.38\ci{9, 14}$ & $22.00\ci{22, 22}$ & $23.06 \ci{22, 24}$ \\
\rowcolor[gray]{0.95} Stop All Cars & $8.56\ci{1, 21}$ & $33.12\ci{21, 41}$ & $7.25 \ci{0, 20}$ \\ 
\rowcolor[gray]{0.95} Start and Stop Cars & $13.50\ci{8, 17}$ & $20.50\ci{20, 21}$ &$8.25\ci{2, 19}$ \\ 
\midrule
\underline{Frostbite} & $2409.38\ci{1668, 3321}$ & $3613.12\ci{3363, 3931}$ & $46.88 \ci{40, 83}$ \\
\rowcolor[gray]{0.95} Static Ice & $41.25\ci{20, 73}$ & $0.00\ci{0, 0}$ & $10.00 \ci{10, 10}$ \\ \midrule
\underline{MsPacman} & $2365.00\ci{2248, 2418}$ & -- & $3174.38 \ci{2792, 3637}$ \\
\rowcolor[gray]{0.95} 2nd Level & $575.62\ci{437, 761}$ & -- & $167.50 \ci{65, 325}$ \\
\rowcolor[gray]{0.95} 3rd Level & $403.12\ci{362, 435}$ & -- & $560.62 \ci{378, 734}$ \\
\rowcolor[gray]{0.95} 4th Level & $216.88\ci{46, 493}$ & -- & $100.62 \ci{75, 120}$ \\ \midrule
\underline{Pong} & $21.00\ci{21, 21}$ & $20.81\ci{20, 21}$ & $21.00 \ci{21, 21}$ \\
\rowcolor[gray]{0.95} Lazy Enemy & $3.44\ci{0, 6}$ & $18.00\ci{16, 19}$ & $-4.38 \ci{-11, -0}$ \\ \midrule
\underline{Riverraid} & $8986.88\ci{8279, 9957}$ & -- & $7849.38 \ci{7434, 8181}$ \\
\rowcolor[gray]{0.95} Color Set 2 & $221.25\ci{180, 319}$ & -- & $392.50 \ci{270, 730}$ \\
\rowcolor[gray]{0.95} Color Set 3 & $241.25\ci{184, 335}$ & -- & $453.12 \ci{280, 678}$ \\
\rowcolor[gray]{0.95} Color Set 4 & $7042.50\ci{6236, 7496}$ & -- & $5759.38 \ci{5137, 6486}$ \\
\rowcolor[gray]{0.95} Linear River & $13066.25\ci{11707, 14301}$ & -- & $6689.38 \ci{5086, 7879}$ \\ \midrule
\underline{SpaceInvaders} & $1035.00\ci{827, 1231}$ & -- & $543.12 \ci{518, 575}$ \\
\rowcolor[gray]{0.95} Shields off by 1 & $1050.94\ci{972, 1150}$ & -- & $705.94 \ci{620, 771}$ \\
\rowcolor[gray]{0.95} Shields off by 3 & $843.12\ci{594, 1152}$ & -- & $578.75 \ci{502, 674}$ \\ \bottomrule
\end{tabular}

    }
    \label{tab:results}
\end{table}

\clearpage
The detailed performance results of PPO across different input representations are provided in \autoref{tab:results-ppo-appendix}. This table presents the raw episodic rewards obtained in both standard (underlined) and perturbed (gray-shaded) environments, averaged over three random seeds. These scores serve as the basis for the normalized performance visualizations in Figures \autoref{fig:ppo_performance} and \autoref{fig:gns}. The results highlight the differences in robustness across input representations, particularly under visual and structural perturbations, offering deeper insight into the trends observed in the main text.

\begin{table}[h]
    \centering
    \setlength{\tabcolsep}{4pt}  
    \renewcommand{\arraystretch}{1.1}  
    
    \caption{This table compares the average episodic rewards (interquartile mean + 95\% confidence interval) achieved by PPO using different input representations: the standard DQN-like representation, OCAtari's Semantic Vector, and our structured masking approaches, including Object Masks, Binary Masks, Class Masks, and Planes. Results are averaged over three random seeds, with standard deviations indicating variability. All agents were trained in the original game environment and evaluated in the standard setting (\underline{underlined}) and visually or behaviorally perturbed variations (gray-shaded). The table highlights how different levels of abstraction in input representations impact the agents' performance and robustness across diverse environments. These results are then normalized and visualized in \autoref{fig:ppo_performance} and \autoref{fig:gns}.}
    \resizebox{\linewidth}{!}{
   \begin{tabular}{lcccccc}
    \toprule
        Game (Variant) & DQN-like & Object Masks & Binary Masks & Class Masks & Planes & Semantic Vector \\
    \midrule
\underline{Amidar} & $564.75 \ci{540, 575}$ & $491.31 \ci{430, 571}$ & $519.75 \ci{413, 618}$ & $448.75 \ci{408, 489}$ & $526.25 \ci{508, 543}$ & $198.81 \ci{153, 231}$ \\
\rowcolor[gray]{0.95} Enemy to Pig & $555.00 \ci{543, 562}$ & $351.94 \ci{294, 430}$ & $530.56 \ci{417, 629}$ & $525.50 \ci{505, 557}$ & $108.00 \ci{91, 118}$ & $322.38 \ci{64, 718}$ \\
\rowcolor[gray]{0.95} Player to Roller & $149.75 \ci{124, 178}$ & $322.50 \ci{316, 332}$ & $540.88 \ci{421, 642}$ & $439.06 \ci{396, 484}$ & $527.12 \ci{509, 544}$ & $202.19 \ci{158, 232}$ \\
\midrule
\underline{BankHeist} & $1176.88 \ci{1139, 1212}$ & $875.62 \ci{855, 892}$ & $1330.00 \ci{1313, 1339}$ & $1300.00 \ci{1240, 1333}$ & $1273.75 \ci{1088, 1446}$ & -- \\
\rowcolor[gray]{0.95} Random City & $1165.00 \ci{1131, 1198}$ & $860.62 \ci{838, 884}$ & $1325.00 \ci{1302, 1342}$ & $1334.38 \ci{1307, 1348}$ & $1275.00 \ci{1086, 1444}$ & -- \\
\midrule
\underline{Bowling} & $67.00 \ci{65, 70}$ & $65.81 \ci{64, 68}$ & $62.62 \ci{60, 65}$ & $69.38 \ci{66, 70}$ & $63.25 \ci{60, 66}$ & $61.88 \ci{60, 65}$ \\
\rowcolor[gray]{0.95} Shift Player & $66.94 \ci{64, 70}$ & $66.88 \ci{64, 69}$ & $64.06 \ci{62, 67}$ & $66.38 \ci{63, 69}$ & $63.31 \ci{60, 67}$ & $61.88 \ci{60, 66}$ \\
\midrule
\underline{Boxing} & $96.31 \ci{95, 98}$ & $97.06 \ci{95, 99}$ & $98.56 \ci{97, 99}$ & $97.44 \ci{96, 98}$ & $99.12 \ci{98, 100}$ & $100.00 \ci{100, 100}$ \\
\rowcolor[gray]{0.95} Red Player, Blue Enemy & $-2.38 \ci{-10, 5}$ & $76.75 \ci{67, 84}$ & $96.19 \ci{95, 98}$ & $96.94 \ci{95, 98}$ & $98.50 \ci{97, 100}$ & $100.00 \ci{100, 100}$ \\
\midrule
\underline{Breakout} & $290.25 \ci{259, 307}$ & $323.75 \ci{283, 364}$ & $360.00 \ci{332, 373}$ & $270.62 \ci{244, 303}$ & $379.69 \ci{364, 396}$ & $48.81 \ci{45, 57}$ \\
\rowcolor[gray]{0.95} All Blocks Red & $306.31 \ci{277, 332}$ & $314.56 \ci{268, 345}$ & $347.19 \ci{317, 367}$ & $279.25 \ci{254, 307}$ & $374.12 \ci{361, 388}$ & $46.94 \ci{44, 56}$ \\
\rowcolor[gray]{0.95} Player and Ball Red & $52.19 \ci{27, 89}$ & $170.50 \ci{105, 251}$ & $348.62 \ci{314, 374}$ & $289.19 \ci{262, 320}$ & $373.81 \ci{360, 388}$  & $47.19 \ci{43, 51}$ \\
\midrule
\underline{Freeway} & $31.75 \ci{30, 32}$ & $33.06 \ci{32, 34}$ & $33.00 \ci{33, 33}$ & $32.44 \ci{32, 33}$ & $33.19 \ci{33, 34}$ & $30.38 \ci{30, 31}$ \\
\rowcolor[gray]{0.95} All Cars Black & $23.06 \ci{22, 24}$ & $22.44 \ci{20, 24}$ & $33.38 \ci{33, 34}$ & $32.56 \ci{32, 33}$ & $33.31 \ci{33, 34}$ & $30.25 \ci{30, 31}$ \\
\rowcolor[gray]{0.95} Stop All Cars & $7.25 \ci{0, 20}$ & $0.00 \ci{0, 0}$ & $0.00 \ci{0, 0}$ & $5.12 \ci{0, 17}$ & $7.81 \ci{0, 20}$ & $0.00 \ci{0, 0}$ \\
\rowcolor[gray]{0.95} Start and Stop Cars & $11.94 \ci{11, 13}$ & $10.81 \ci{10, 12}$ & $10.56 \ci{9, 12}$ & $12.62 \ci{11, 14}$ & $12.88 \ci{11, 14}$ & $7.06 \ci{5, 9}$ \\
\midrule
\underline{Frostbite} & $46.88 \ci{40, 83}$ & $46.88 \ci{40, 79}$ & $46.88 \ci{40, 79}$ & $51.88 \ci{40, 74}$ & $48.75 \ci{40, 81}$ & $62.50 \ci{47, 82}$ \\
\rowcolor[gray]{0.95} Static Ice & $10.00 \ci{10, 10}$ & $0.00 \ci{0, 1}$ & $5.00 \ci{2, 9}$ & $0.00 \ci{0, 0}$ & $2.50 \ci{0, 17}$ & $2.50 \ci{0, 19}$ \\
\midrule
\underline{MsPacman} & $3174.38 \ci{2792, 3637}$ & $5880.00 \ci{5531, 6068}$ & $4833.12 \ci{4583, 5221}$ & $4549.38 \ci{3810, 5188}$ & $7187.50 \ci{6779, 7415}$ & $1696.88 \ci{1268, 2471}$ \\
\rowcolor[gray]{0.95} 2nd Level & $167.50 \ci{65, 325}$ & $435.00 \ci{305, 635}$ & $426.25 \ci{354, 575}$ & $384.38 \ci{245, 560}$ & $263.75 \ci{189, 357}$ & $72.50 \ci{60, 80}$ \\
\rowcolor[gray]{0.95} 3rd Level & $560.62 \ci{378, 734}$ & $310.00 \ci{276, 338}$ & $331.25 \ci{262, 409}$ & $418.75 \ci{322, 634}$ & $120.00 \ci{100, 164}$ & $109.38 \ci{56, 183}$ \\
\rowcolor[gray]{0.95} 4th Level & $100.62 \ci{75, 120}$ & $253.75 \ci{210, 288}$ & $663.75 \ci{592, 751}$ & $491.25 \ci{184, 824}$ & $137.50 \ci{93, 204}$ & $99.38 \ci{54, 171}$ \\
\midrule
\underline{Pong} & $21.00 \ci{21, 21}$ & $21.00 \ci{21, 21}$ & $21.00 \ci{21, 21}$ & $21.00 \ci{21, 21}$ & $21.00 \ci{21, 21}$ & $21.00 \ci{21, 21}$ \\
\rowcolor[gray]{0.95} Lazy Enemy & $-4.38 \ci{-11, -0}$ & $20.00 \ci{20, 20}$ & $11.25 \ci{-2, 20}$ & $8.25 \ci{3, 14}$ & $12.44 \ci{-1, 20}$ & $-21.00 \ci{-21, -21}$ \\
\midrule
\underline{Riverraid} & $7849.38 \ci{7434, 8181}$ & $7938.75 \ci{7780, 8068}$ & $8076.88 \ci{7872, 8351}$ & $8160.00 \ci{7998, 8274}$ & $7953.75 \ci{7841, 8188}$ & $3933.75 \ci{3554, 4273}$ \\
\rowcolor[gray]{0.95} Color Set 2 & $392.50 \ci{270, 730}$ & $7674.38 \ci{7404, 7874}$ & $8178.75 \ci{7936, 8558}$ & $8164.38 \ci{7988, 8274}$ & $7995.00 \ci{7872, 8138}$ & $3583.12 \ci{3159, 3955}$ \\
\rowcolor[gray]{0.95} Color Set 3 & $453.12 \ci{280, 678}$ & $7687.50 \ci{7512, 7878}$ & $8225.62 \ci{7978, 8593}$ & $8094.38 \ci{7941, 8245}$ & $8060.62 \ci{7878, 8298}$ & $3813.75 \ci{3467, 4092}$ \\
\rowcolor[gray]{0.95} Color Set 4 & $5759.38 \ci{5137, 6486}$ & $7435.00 \ci{7279, 7638}$ & $8199.38 \ci{7961, 8481}$ & $8065.62 \ci{7909, 8223}$ & $7961.25 \ci{7824, 8188}$ & $3632.50 \ci{3212, 3996}$ \\
\rowcolor[gray]{0.95} Linear River & $6689.38 \ci{5086, 7879}$ & $2903.75 \ci{2750, 3179}$ & $2705.00 \ci{2690, 2724}$ & $2832.50 \ci{2680, 3080}$ & $2931.88 \ci{2761, 3172}$ & $3062.50 \ci{2952, 3162}$ \\
\midrule
\underline{SpaceInvaders} & $543.12 \ci{518, 575}$ & $905.94 \ci{769, 967}$ & $911.56 \ci{797, 1008}$ & $801.88 \ci{703, 896}$ & $988.12 \ci{850, 1271}$ & $377.81 \ci{315, 438}$ \\
\rowcolor[gray]{0.95} Shields off by 1 & $705.94 \ci{620, 771}$ & $761.25 \ci{592, 909}$ & $779.38 \ci{738, 823}$ & $703.12 \ci{512, 811}$ & $2170.62 \ci{1676, 2697}$ & $177.19 \ci{155, 198}$ \\
\rowcolor[gray]{0.95} Shields off by 3 & $578.75 \ci{502, 674}$ & $484.06 \ci{380, 592}$ & $705.31 \ci{593, 819}$ & $317.81 \ci{226, 415}$ & $798.12 \ci{591, 1012}$ & $397.50 \ci{285, 545}$ \\
    \bottomrule
    \end{tabular}}
    \label{tab:results-ppo-appendix}
\end{table}

\begin{figure}[h]
    \centering
    \includegraphics[width=\linewidth]{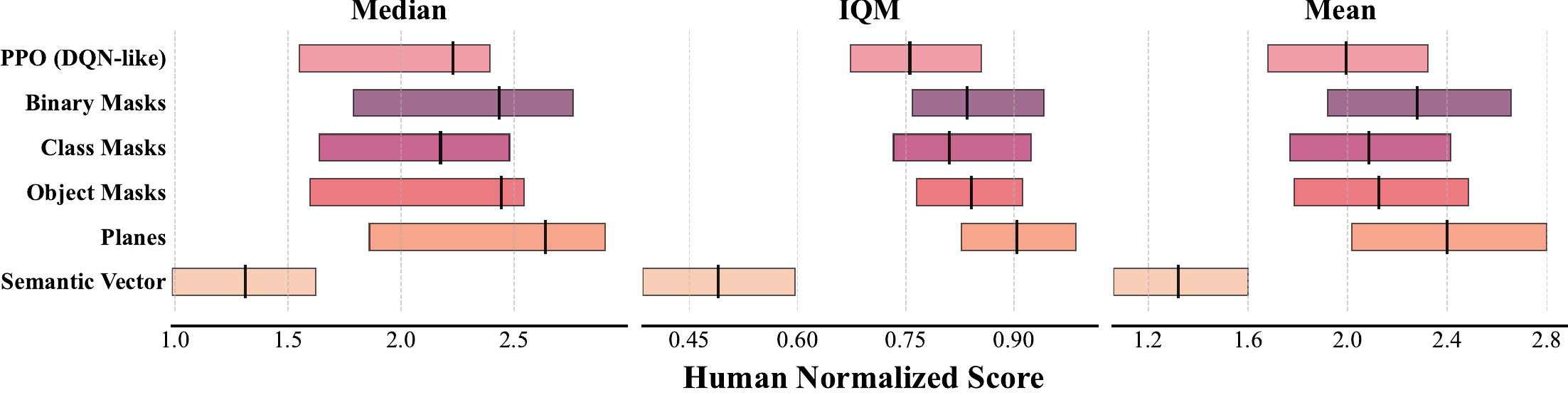}
    \caption{Extended Version of \autoref{fig:ppo_performance}. Next to the interquartile mean (IQM), we also report the Median and Mean. The main message that OCCAM can keep up with PPO also can be seen in these metrics. Numbers are from \autoref{tab:results-ppo-appendix}.}
    \label{fig:visuals-extended}
\end{figure}

\begin{figure}[h]
    \centering
    \includegraphics[width=\linewidth]{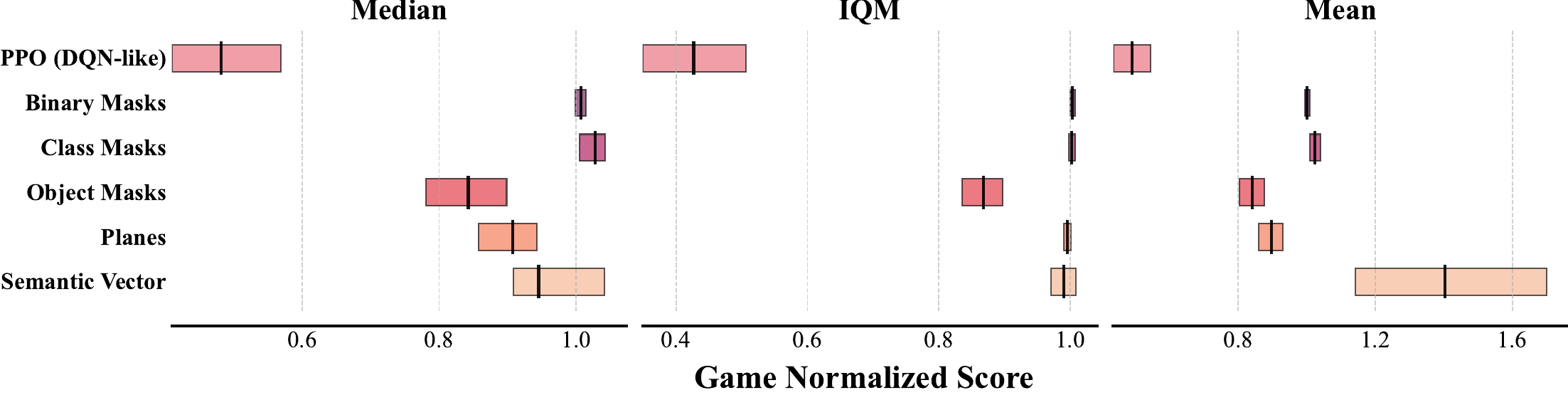}
    \caption{Extended Version of \autoref{fig:gns}a: Visual changes. Next to the interquartile mean (IQM), we also report the Median and Mean. The main message that OCCAM can mitigate performance loss in visually altered environments can also be seen in these metrics. Numbers are from \autoref{tab:results-ppo-appendix}.}
    \label{fig:logic-extended}
\end{figure}

\begin{figure}[h]
    \centering
    \includegraphics[width=\linewidth]{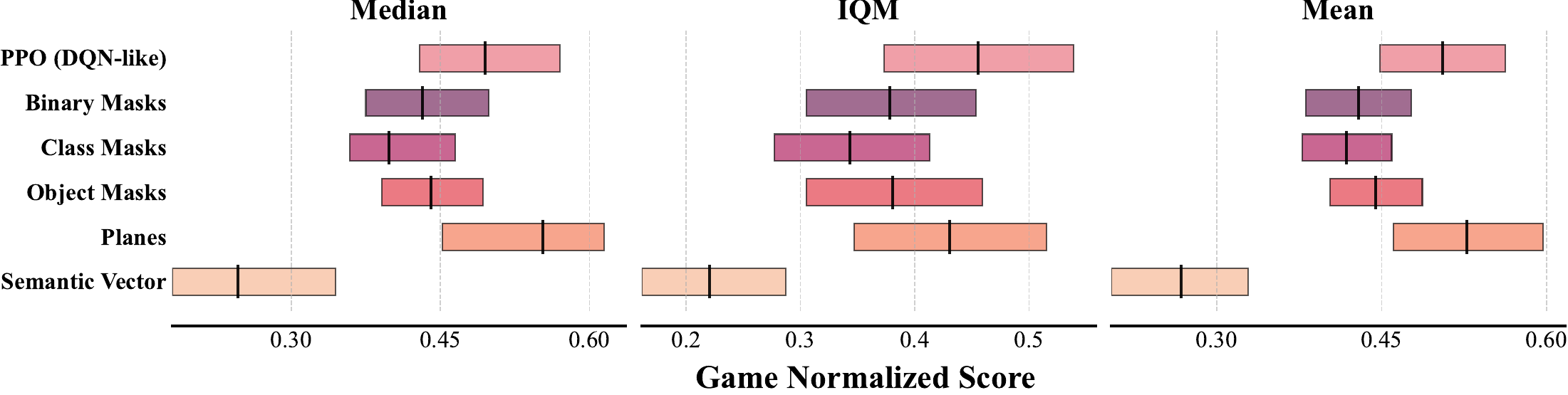}
    \caption{Extended Version of \autoref{fig:gns}b: Game logic changes. Next to the interquartile mean (IQM), we also report the Median and Mean. These metrics also show that OCCAM approaches still remain vulnerable to game logic changes. Numbers are from \autoref{tab:results-ppo-appendix}.}
    \label{fig:enter-label}
\end{figure}

\begin{sidewaystable}[h]
    \centering
    \setlength{\tabcolsep}{4pt}
    \renewcommand{\arraystretch}{1.4}
    \caption{\textbf{Evaluation with \textit{Repeat Action Probability} = 0.25}.  These results extend Tables \ref{tab:results} and \ref{tab:results-ppo-appendix} by adding a robustness test. Enabling \textit{"Sticky Actions"}, as proposed by \citet{MachadoBTVHB18}, is a common practice in evaluating Atari agents and is provided to make comparison to related work easier.}
    \resizebox{\linewidth}{!}{
    \begin{tabular}{lccccccccc}
    \toprule
        Game (Variant) & DQN & C51 & MDQN & PPO & Object Masks & Binary Masks & Class Masks & Planes & Semantic Vector \\
    \midrule
\underline{Amidar} & $431.44 \ci{318, 575}$ &  --  & $749.56 \ci{698, 810}$ & $566.25 \ci{545, 576}$ & $553.69 \ci{493, 615}$ & $525.12 \ci{430, 605}$ & $478.88 \ci{442, 513}$ & $527.06 \ci{509, 552}$ & $217.88 \ci{186, 241}$ \\
\rowcolor[gray]{0.95} Enemy to Pig & $425.31 \ci{352, 549}$ &  --  & $675.56 \ci{518, 838}$ & $555.31 \ci{542, 565}$ & $383.56 \ci{332, 426}$ & $512.69 \ci{415, 597}$ & $516.12 \ci{465, 554}$ & $107.75 \ci{91, 118}$ & $412.12 \ci{129, 795}$ \\
\rowcolor[gray]{0.95} Player to Roller & $60.44 \ci{50, 77}$ &  --  & $91.94 \ci{75, 112}$ & $162.50 \ci{132, 197}$ & $314.88 \ci{251, 369}$ & $534.56 \ci{435, 610}$ & $492.31 \ci{458, 526}$ & $509.50 \ci{491, 528}$ & $206.25 \ci{168, 237}$ \\
\midrule
\underline{BankHeist} & $1076.25 \ci{1007, 1132}$ &  --  & $1406.88 \ci{1302, 1478}$ & $1043.75 \ci{1004, 1084}$ & $780.62 \ci{769, 794}$ & $1191.88 \ci{1158, 1211}$ & $1181.25 \ci{1151, 1205}$ & $1154.38 \ci{1013, 1302}$ &  --  \\
\rowcolor[gray]{0.95} Random City & $1065.00 \ci{991, 1135}$ &  --  & $1378.75 \ci{1282, 1460}$ & $1043.75 \ci{1006, 1075}$ & $787.50 \ci{761, 805}$ & $1166.25 \ci{1119, 1202}$ & $1181.25 \ci{1139, 1204}$ & $1153.12 \ci{1008, 1314}$ &  --  \\
\midrule
\underline{Bowling} & $34.38 \ci{30, 37}$ & $30.38 \ci{24, 38}$ & $31.94 \ci{30, 34}$ & $66.31 \ci{64, 68}$ & $65.44 \ci{63, 67}$ & $64.75 \ci{62, 67}$ & $69.94 \ci{67, 71}$ & $63.94 \ci{61, 67}$ & $61.88 \ci{60, 65}$ \\
\rowcolor[gray]{0.95} Shift Player & $32.69 \ci{25, 37}$ & $39.50 \ci{34, 43}$ & $29.12 \ci{27, 33}$ & $64.38 \ci{62, 67}$ & $67.56 \ci{64, 70}$ & $63.25 \ci{61, 66}$ & $67.19 \ci{64, 70}$ & $62.44 \ci{60, 66}$ & $62.12 \ci{60, 65}$ \\
\midrule
\underline{Boxing} & $88.12 \ci{84, 92}$ & $74.38 \ci{68, 80}$ & $92.19 \ci{89, 94}$ & $93.06 \ci{91, 95}$ & $94.38 \ci{92, 96}$ & $96.06 \ci{95, 97}$ & $94.44 \ci{93, 96}$ & $97.38 \ci{96, 98}$ & $91.19 \ci{88, 94}$ \\
\rowcolor[gray]{0.95} Red Player, Blue Enemy & $0.38 \ci{-2, 3}$ & $-1.88 \ci{-5, 1}$ & $1.06 \ci{-2, 3}$ & $8.69 \ci{4, 14}$ & $79.94 \ci{76, 83}$ & $95.75 \ci{95, 97}$ & $94.56 \ci{93, 96}$ & $96.25 \ci{95, 98}$ & $92.06 \ci{88, 95}$ \\
\midrule
\underline{Breakout} & $69.25 \ci{51, 93}$ & $35.12 \ci{28, 49}$ & $96.44 \ci{71, 141}$ & $148.25 \ci{117, 191}$ & $216.06 \ci{156, 279}$ & $258.56 \ci{212, 295}$ & $222.19 \ci{171, 267}$ & $371.12 \ci{348, 390}$ & $32.81 \ci{28, 38}$ \\
\rowcolor[gray]{0.95} All Blocks Red & $235.94 \ci{169, 297}$ & $34.38 \ci{24, 49}$ & $234.25 \ci{167, 305}$ & $175.38 \ci{119, 236}$ & $298.69 \ci{247, 328}$ & $278.19 \ci{229, 312}$ & $259.19 \ci{207, 290}$ & $372.12 \ci{330, 403}$ & $38.88 \ci{32, 45}$ \\
\rowcolor[gray]{0.95} Player and Ball Red & $7.88 \ci{6, 11}$ & $5.00 \ci{3, 7}$ & $24.00 \ci{19, 30}$ & $36.12 \ci{19, 60}$ & $99.19 \ci{72, 155}$ & $255.81 \ci{201, 287}$ & $216.12 \ci{165, 266}$ & $390.12 \ci{352, 407}$ & $34.19 \ci{29, 40}$ \\
\midrule
\underline{Freeway} & $26.44 \ci{16, 33}$ & $32.81 \ci{32, 33}$ & $33.25 \ci{33, 34}$ & $31.38 \ci{30, 32}$ & $32.56 \ci{32, 33}$ & $33.00 \ci{33, 33}$ & $32.12 \ci{32, 32}$ & $33.10 \ci{33, 34}$ & $30.69 \ci{30, 31}$ \\
\rowcolor[gray]{0.95} All Cars Black & $11.81 \ci{9, 14}$ & $21.31 \ci{21, 22}$ & $24.44 \ci{24, 25}$ & $23.12 \ci{22, 24}$ & $23.19 \ci{21, 25}$ & $32.62 \ci{32, 33}$ & $32.31 \ci{32, 33}$ & $33.30 \ci{33, 34}$ & $30.44 \ci{30, 31}$ \\
\rowcolor[gray]{0.95} Stop All Cars & $8.81 \ci{1, 21}$ & $38.50 \ci{36, 40}$ & $40.00 \ci{40, 40}$ & $7.50 \ci{0, 20}$ & $0.00 \ci{0, 0}$ & $0.00 \ci{0, 0}$ & $7.12 \ci{0, 19}$ & $21.90 \ci{6, 37}$ & $0.00 \ci{0, 0}$ \\
\rowcolor[gray]{0.95} Start and Stop Cars & $13.75 \ci{8, 18}$ & $21.19 \ci{21, 22}$ & $21.88 \ci{21, 23}$ & $20.50 \ci{20, 22}$ & $18.50 \ci{17, 20}$ & $18.88 \ci{18, 20}$ & $19.06 \ci{18, 20}$ & $20.40 \ci{20, 21}$ & $14.44 \ci{12, 16}$ \\
\midrule
\underline{Frostbite} & $3273.75 \ci{2649, 3838}$ & $3445.62 \ci{3063, 3713}$ & $4386.25 \ci{2546, 5908}$ & $295.62 \ci{287, 303}$ & $277.50 \ci{270, 290}$ & $290.00 \ci{278, 304}$ & $265.00 \ci{258, 270}$ & $277.50 \ci{271, 286}$ & $257.50 \ci{251, 325}$ \\
\rowcolor[gray]{0.95} Static Ice & $38.75 \ci{19, 70}$ & $0.00 \ci{0, 0}$ & $28.12 \ci{10, 64}$ & $40.00 \ci{40, 49}$ & $10.62 \ci{0, 32}$ & $3.75 \ci{0, 11}$ & $1.25 \ci{0, 24}$ & $28.12 \ci{0, 79}$ & $30.00 \ci{0, 83}$ \\
\midrule
\underline{MsPacman} & $2332.50 \ci{2164, 2497}$ &  --  & $2406.25 \ci{2252, 2586}$ & $3172.50 \ci{2844, 3622}$ & $5195.00 \ci{4476, 5724}$ & $4226.88 \ci{3646, 4896}$ & $4313.12 \ci{3790, 4899}$ & $6211.25 \ci{5471, 6794}$ & $1546.88 \ci{1334, 2239}$ \\
\rowcolor[gray]{0.95} 2nd Level & $486.25 \ci{379, 627}$ &  --  & $334.38 \ci{294, 384}$ & $146.25 \ci{72, 294}$ & $280.62 \ci{236, 319}$ & $547.50 \ci{400, 778}$ & $377.50 \ci{256, 553}$ & $293.75 \ci{234, 366}$ & $72.50 \ci{60, 80}$ \\
\rowcolor[gray]{0.95} 3rd Level & $370.00 \ci{322, 429}$ &  --  & $389.38 \ci{337, 463}$ & $476.25 \ci{362, 578}$ & $285.00 \ci{258, 317}$ & $376.88 \ci{322, 439}$ & $695.00 \ci{447, 969}$ & $126.88 \ci{104, 161}$ & $95.62 \ci{55, 169}$ \\
\rowcolor[gray]{0.95} 4th Level & $216.25 \ci{48, 490}$ &  --  & $195.62 \ci{148, 248}$ & $100.62 \ci{77, 118}$ & $259.38 \ci{222, 301}$ & $628.75 \ci{512, 774}$ & $228.12 \ci{110, 501}$ & $124.38 \ci{98, 172}$ & $80.00 \ci{51, 139}$ \\
\midrule
\underline{Pong} & $16.56 \ci{16, 17}$ & $10.75 \ci{9, 12}$ & $17.44 \ci{16, 18}$ & $16.88 \ci{16, 18}$ & $18.25 \ci{18, 19}$ & $19.00 \ci{18, 20}$ & $19.25 \ci{19, 20}$ & $19.31 \ci{19, 20}$ & $18.75 \ci{18, 19}$ \\
\rowcolor[gray]{0.95} Lazy Enemy & $-3.94 \ci{-7, -1}$ & $1.25 \ci{-3, 5}$ & $-6.94 \ci{-10, -3}$ & $-7.06 \ci{-10, -4}$ & $12.06 \ci{10, 15}$ & $11.56 \ci{5, 16}$ & $2.62 \ci{-4, 9}$ & $16.88 \ci{16, 18}$ & $-19.25 \ci{-21, -16}$ \\
\midrule
\underline{Riverraid} & $9423.12 \ci{8740, 10265}$ &  --  & $8400.62 \ci{8051, 9074}$ & $7888.12 \ci{7611, 8086}$ & $7947.50 \ci{7814, 8092}$ & $7957.50 \ci{7739, 8204}$ & $7803.12 \ci{7552, 7993}$ & $7990.00 \ci{7824, 8198}$ & $3285.00 \ci{2934, 3674}$ \\
\rowcolor[gray]{0.95} Color Set 2 & $245.00 \ci{188, 376}$ &  --  & $898.12 \ci{766, 1071}$ & $413.75 \ci{278, 771}$ & $7637.50 \ci{7474, 7811}$ & $7850.00 \ci{7745, 8011}$ & $7866.88 \ci{7567, 8105}$ & $8041.88 \ci{7841, 8242}$ & $3304.38 \ci{3021, 3499}$ \\
\rowcolor[gray]{0.95} Color Set 3 & $260.00 \ci{192, 370}$ &  --  & $835.00 \ci{576, 1062}$ & $493.75 \ci{355, 606}$ & $7610.62 \ci{7368, 7851}$ & $7940.00 \ci{7732, 8173}$ & $7672.50 \ci{7407, 7982}$ & $8216.88 \ci{8002, 8436}$ & $3418.75 \ci{3037, 3755}$ \\
\rowcolor[gray]{0.95} Color Set 4 & $6350.00 \ci{5508, 7023}$ &  --  & $5135.62 \ci{4576, 5924}$ & $5541.88 \ci{4646, 6094}$ & $7450.62 \ci{7339, 7626}$ & $7903.12 \ci{7677, 8154}$ & $7855.00 \ci{7646, 8042}$ & $8250.62 \ci{8048, 8462}$ & $3426.88 \ci{3123, 3814}$ \\
\rowcolor[gray]{0.95} Linear River & $12373.12 \ci{11256, 13166}$ &  --  & $12165.62 \ci{11121, 13146}$ & $8103.75 \ci{7804, 8843}$ & $2909.38 \ci{2750, 3160}$ & $2702.50 \ci{2681, 2724}$ & $2885.00 \ci{2720, 3156}$ & $2882.50 \ci{2718, 3152}$ & $3111.25 \ci{2954, 3289}$ \\
\midrule
\underline{SpaceInvaders} & $1054.06 \ci{820, 1342}$ &  --  & $2162.50 \ci{1781, 2484}$ & $745.31 \ci{688, 792}$ & $688.44 \ci{600, 806}$ & $826.25 \ci{763, 909}$ & $554.06 \ci{475, 634}$ & $1556.88 \ci{1247, 1851}$ & $377.19 \ci{301, 455}$ \\
\rowcolor[gray]{0.95} Shields off by 1 & $1001.56 \ci{812, 1173}$ &  --  & $1538.75 \ci{1153, 1874}$ & $694.69 \ci{620, 778}$ & $621.25 \ci{560, 693}$ & $802.81 \ci{730, 884}$ & $483.12 \ci{362, 588}$ & $1786.56 \ci{1416, 2217}$ & $294.69 \ci{260, 332}$ \\
\rowcolor[gray]{0.95} Shields off by 3 & $606.25 \ci{509, 720}$ &  --  & $812.19 \ci{646, 1013}$ & $677.19 \ci{572, 761}$ & $468.75 \ci{389, 571}$ & $589.38 \ci{474, 683}$ & $287.81 \ci{245, 325}$ & $1555.00 \ci{1189, 1899}$ & $379.69 \ci{328, 461}$ \\
\midrule
    \bottomrule
    \end{tabular}}
\end{sidewaystable}

\clearpage
\subsection{Reward Function Adjustment in Skiing}
\label{app:skiing}

The standard reward function in Skiing is known to be malformed and sparse, making it difficult for reinforcement learning agents to optimize their behavior effectively (see \citet{delfosse2024interpretable}). To enable training, we modified the reward function to incorporate orientation, speed, and score changes, allowing agents to learn meaningful policies. However, all final evaluations use the original reward function to ensure comparability.

The reward function is 

\begin{minted}[fontsize=\small, linenos, breaklines, tabsize=4]{python}
REWARD = 32

def reward_function(self) -> float:
    global REWARD
    ram = self.get_ram()
    orientation = -abs(8 - ram[15]) * 0.5
    speed = ram[14] * 0.01
    score = ram[107]
    reward = orientation + speed

    if score != REWARD:
        reward += 100000

    REWARD = ram[107]
    return reward
\end{minted}

\end{document}